\documentclass[11pt]{article}

\usepackage[english]{babel}

\usepackage[letterpaper,margin=1in]{geometry}
\usepackage{makecell}
\usepackage{multirow}
\usepackage{amsmath}
\usepackage{bm}
\usepackage{url}
\usepackage{graphicx}
\usepackage{amsfonts}
\usepackage{csquotes}
\usepackage{booktabs}
\usepackage{floatrow}
\usepackage{subcaption}
\newfloatcommand{capbtabbox}{table}[][\FBwidth]
\usepackage[colorlinks=true, allcolors=blue]{hyperref}

\newcommand{\cO}{\mathcal{O}}

\newcommand{\setfoot}[2]{%
    \footnote{#2}%
    \newcounter{#1}%
    \setcounter{#1}{\value{footnote}}%
}

\newcommand{\getfoot}[1]{%
    \footnotemark[\value{#1}]%
}

\title{Probabilistic Spatiotemporal Modeling of Day-Ahead Wind Power Generation with Input-Warped Gaussian Processes}
\author{Qiqi Li\setfoot{unia}{Department of Statistics and Applied Probability, University of California Santa Barbara, Santa Barbara, CA USA; email: \{qiqili, ludkovski\}@pstat.ucsb.edu;}
~\setfoot{unib}{The work presented here was done prior to joining Amazon and does not relate to this author’s position at Amazon.}  \and Michael Ludkovski\getfoot{unia}}

\begin{document}
\maketitle

\begin{abstract}
We design a Gaussian Process (GP) spatiotemporal model to capture features of day-ahead wind power forecasts. We work with hourly-scale day-ahead forecasts across hundreds of wind farm locations, with the main aim of constructing a fully probabilistic joint model across space and hours of the day. To this end, we design a separable space-time kernel, implementing both temporal and spatial input warping to capture the non-stationarity in the covariance of wind power. We conduct synthetic experiments to validate our choice of the spatial kernel and to demonstrate the effectiveness of warping in addressing nonstationarity. The second half of the paper is devoted to a detailed case study using a realistic, fully calibrated dataset representing wind farms in the ERCOT region of Texas.

\vspace{2em}

\end{abstract}

\noindent
\section{Introduction}

Wind power is one of the fastest-growing renewable energy sectors and a key pillar for the transition to a carbon-free economy. In 2023, energy from wind accounted for 10.2\% of all U.S.~utility-scale electricity generation \cite{energysource}. Being intrinsically weather-driven, wind power injects uncertainty into the balancing of power demand and generation. On the daily operational time scale, quantifying the asset-specific and area-wide uncertainty of renewable generation for the next day is an essential ingredient of grid management. Specifically, grid operators need probabilistic spatiotemporal forecasting of wind power in order to appropriately set grid reserves, ensure grid stability, and optimize dispatch of grid resources.

Our goal is to develop a statistical framework for short-term wind power generation simulations across space and time. This project is motivated by working with a large dataset of wind generation in the Electric Reliability Council of Texas (ERCOT) region
and is geared to the concrete practical concerns faced by electricity grid operators. We refer to our team's related publications \cite{carmona2023joint,carmona2024cost,terren2023extreme,ludkovski2022} that employ similar simulations for various downstream risk management tasks; other use cases are discussed, among others, in \cite{haupt2019use,lara2021multi,li2020review,wang2016quantifying}.

The motivating dataset, described in more detail in Sections \ref{sec: eda} and  \ref{sec: data_analysis},  has been recently created by several other teams as part of the ARPA-E PERFORM Data Plan suite \cite{arpae2024data}. Its setting is tailored to the purposes of daily electric grid operations and implies several notable differences compared to standard spatiotemporal statistical setups. First, it directly reports hourly power generation in megawatt-hours (MWh), rather than wind speed. Hence, it operates in energy units rather than meteorological quantities. Second, the dataset provides not only the realized generation but also respective day-ahead point forecasts. These forecasts are produced daily and make the paired forecast-actuals no longer a time-series but a sequence of daily blocks. This pairing reflects the reality that the primary grid uncertainty quantification is conditional on the latest available weather forecast and is done once a day. Third, the provided measurements are hourly (with no missing dates), and are only available at wind farm locations, forming a highly irregular and non-uniform spatial pattern. 

The above dataset features bring forth the following statistical tasks that our analysis resolves. Firstly, given the forecast-actuals pairing, we focus on modeling the respective forecast errors. In turn, this makes our data effectively independent across days, 
so that we treat the dataset as an i.i.d.~sample of 24-long vectors. The conditioning on the external forecast moreover implies that our main goal is not prediction per se (already available by the forecast itself), but uncertainty quantification, especially across multiple spatial locations. Indeed, the fundamental task faced by grid operators is to properly anticipate likely forecast deviations across (sub-regions of) the grid and set aside sufficient power reserves, which hinges on building a calibrated joint area-wide model. Spatial uncertainty quantification calls for accurate modeling of the respective correlation structure, in particular paying attention to respective non-stationarity due to the underlying physical geography features.

In the power engineering community, wind power generation is modeled using physical approaches like numerical weather prediction \cite{al2010review} frameworks that solve complex mathematical models for atmospheric conditions. Since the underlying models integrate weather data and physical descriptions,  these methods have very high computational costs and varying precision depending on weather conditions \cite{potter2006very}, making them less suited for short-term forecasting. As a result, there has been a trend towards statistical approaches that utilize machine learning techniques to learn pattern directly from the data without reference to a physical systems  \cite{giebel2017wind}. Existing techniques for short-term (hourly-scale) forecasting include neural networks \cite{dong2021spatio,medina2020performance,potter2006very}, Gaussian processes \cite{chen2013short}, ensemble models \cite{lee2020wind}, Markovian graphs \cite{lara2021multi} copulas \cite{ludkovski2022}, transformed autoregressive time series \cite{pinson2009probabilistic} and support vector machines \cite{yuan2015short}. Two older surveys are in \cite{giebel2017wind,soman2010review}. 

In the statistical community, most of the existing relevant literature concentrates on modeling wind \emph{speed}. Existing methods include regime-switching space-time (RST) models \cite{gneiting2006calibrated,jia2022short}, \cite{zhu2012short},  neural networks (NN) \cite{azad2014long} and stochastic differential equations \cite{iversen2016short} and typically incorporate meteorological factors, such as atmospheric pressure, temperature, and hub height.
Within the operating range of the wind turbine, wind power is roughly proportional to the cube of wind speed \cite{energyedu2024windpower}, implying that small errors in wind speed can significantly impact power predictions. Moreover, in our case we do not have access to operational characteristics like the turbine manufacturer's power curve to accurately convert wind speed into MWh. 

Closely related to wind power modeling is the field of solar power, where the fundamental meteorological variable is solar irradiance. Unlike wind power, solar energy is only produced for a few hours a day and the maximum capacity is hour-dependent. Temporal dependence is driven by cloud cover rather than wind patterns. See \cite{najibi2021enhanced,sheng2017short} for GP-based approaches and \cite{doubleday2020probabilistic,woodruff2018constructing,van2021benchmark,heaton2019case} for other probabilistic forecasts.

The vast majority of extant statistical methods are either geared to a single wind asset or produce a point forecast  \cite{dong2021spatio, medina2020performance, lee2020wind, yuan2015short}, with no quantitative information about the uncertainty inherent in power generation. Thus, the literature on probabilistic area-wide forecasting is much sparser \cite{arrieta2022spatio, tastu2015space, dowell2015very}
and is dominated by Gaussian Process (GP) frameworks or Bayesian deep learning \cite{liu2020probabilistic}. GPs \cite{rasmussen2003gaussian} is one of the most common modeling strategies in spatial statistics \cite{banerjee2008gaussian}, providing a flexible statistical inference framework with inherent uncertainty quantification. It is widely applied to model weather conditions and renewable energy \cite{chen2013short,wang2021use, najibi2021enhanced, sheng2017short, yan2015hybrid}. 
In the closest approach to ours, Chen et al. \cite{chen2013short} utilize wind speed data from an NWP model to feed into GP, focusing on achieving accurate power generation predictions rather than quantifying the associated uncertainties. 

To build a statistical model for wind power generation based solely on spatial and temporal inputs  we adopt the GP paradigm thanks to its flexibility and inherent capability for probabilistic forecasting. 
%
Standard GP modeling assumes  that the covariance kernel is stationary across the input space, which is unrealistic for our application. To address this challenge we adopt the input warping technique as introduced in Zammit et al.~\cite{zammit2022deep} and extended to spatiotemporal setting by Vu et al.~\cite{vu2023constructing}. 
Our data exhibits both spatial and temporal nonstationarity, with mixed-type inputs where space coordinates are continuous and the time domain is discretized.
The spatial domain exhibits nonstationarity due to the natural landscape, while the temporal domain shows a nonstationary and periodic pattern over the 24-hour grid. 
To this end we construct a bespoke separable kernel for this application.

Although our initial dataset contains over a million data points, because we treat it as independent daily samples, we only ever have to deal with matrices of a few thousand rows during our analysis. Consequently, we are able to perform exact inference in our GPs and do not need to resort to large-scale GP approximations that are front and in center in many related contexts \cite{heaton2019case,katzfuss2021general}. We emphasize that our choice is driven not by computational considerations but by the application itself which concentrates on the intra-day spatial modeling.

In sum, given geographical longitude and latitude coordinates along with historical data from multiple wind assets, we build and calibrate a spatiotemporal GP. The resulting day-ahead probabilistic forecast engine generates scenarios whose space-time covariance effectively models the dependencies in observed wind power generation. The produced simulations can be used for conditional analysis given a set of realized wind power data at other locations and/or hours, or more commonly for area-wide simulations to capture uncertainty in generation across the next day based on the point weather forecast. Notably, the model can be used directly to output  uncertainty metrics, such as predictive quantiles, as well as a simulation engine that outputs scenarios, e.g., for stochastic programming.

The rest of the paper is organized as follows. Section~\ref{sec: eda} introduces our Texas wind power dataset and the data preprocessing procedure. Section~\ref{sec: modeling} provides a summary of model details, including a review of spatiotemporal Gaussian Processes, kernel structures and the input warping technique. In Section~\ref{sec: eval} we summarize various evaluation metrics for model predictive performance. Section~\ref{sec: synthetic} discusses synthetic results about kernel choice and warping recovery. Results for the Texas dataset and model-based simulations are in Section~\ref{sec: data_analysis}; Section~\ref{sec: conclusion} concludes.

\section{Data Description}
\label{sec: eda}
\begin{figure}[ht]
  \centering
  \includegraphics[width=0.5\textwidth]{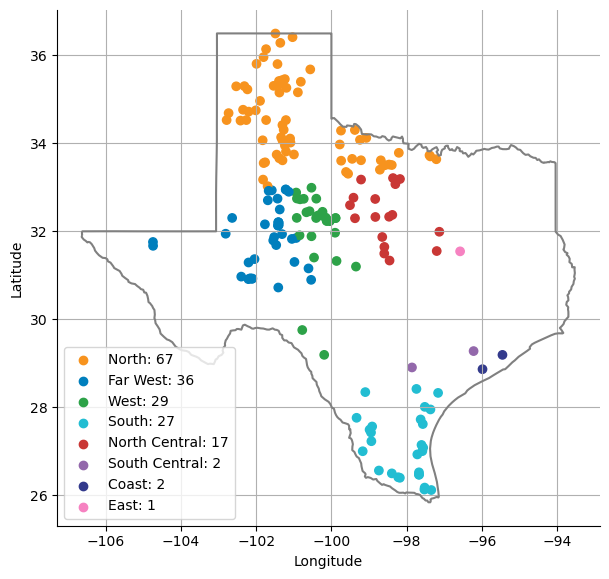}
  \caption{Locations $\mathbf{s}_m$ of the $m=1,\ldots,181$ wind power facilities in a longitude-latitude coordinate map, colored by the 8 ERCOT zones. Most wind farms are in the Northern and Western areas of Texas.}
  \label{fig:locs}
\end{figure}

Our work is based on a unique dataset representing wind power production from the Electric Reliability Council of Texas (ERCOT) control area.  Funded by ARPA-E's PERFORM Program \cite{arpae2024data}, this dataset is a highly realistic and detailed database of grid-wide renewable generation serving as a basis for cutting-edge grid management optimization platforms. The data is synthetically generated and was created by multiple teams based on re-analysis of ECMWF numerical weather prediction system and satellite meteorological imagery. It consists of hourly wind production (in megawatts, MW) during the calendar year 2018 for hundreds of wind farms in ERCOT.
%
ERCOT consists of 8 zones (Coast, West, Far West, North, North Central, East, Southern, and South Central), which are used to delineate the geographic distribution of the wind farms see Figure~\ref{fig:locs}. 
Geographic locations $\mathbf{s}$ are specified as longitude-latitude coordinate pairs, ranging from $26.12$ to $36.50$ degrees latitude, and $-104.74$ to $-95.46$ degrees longitude. Certain wind farms were excluded from the analysis due to either overlapping spatial coordinates or the presence of anomalous data collection.
The dataset includes day-ahead ECMWF-based hourly-scale forecasts $p^F_{m,t,n}$ of wind generation and paired realized actuals $p^A_{m,t,n}$, both indexed by location $\mathbf{s}_m$, hour $j$ and day $n$,  $m \in \{1,2,\ldots,M\}, j\in\{1,2,\ldots,T\}, n\in \{1,2, \ldots,N\}$. The temporal domain is fixed throughout to be $T=24$ hours. For each facility, the day-ahead forecast is generated in batch based on the medium-term ECMWF model. This forecast is issued at noon, hence is 12-36 hours into the future; all the 24 hourly forecast values are issued at once.  One motivation for our work is that while the ARPA-E dataset took significant computational resources (several months of NREL supercomputers) to produce, it contains no quantification of joint uncertainty; instead only marginal predictive standard deviation is reported based on ECMWF ensemble forecasts.

\subsection*{Normalization and Deseasonalization}
The forecasted $p^F$ and realized power generation $p^A$ fall within a finite range bounded below by zero, which occurs when the wind speed is below the turbine's cut-in speed, and the maximum capacity, which is reached when all turbines within the wind farm are operating at their peak power output. Approximately $5.5\%$ of the actual wind power values, $p^A_{m,j,n}$, are exactly zero, while $4.9\%$ reach the maximum capacity of their respective wind assets.

Wind power production displays considerable temporal seasonality, fluctuating annually and daily.  In general in ERCOT, wind tends to be highest during the spring and lowest during the mid-to-late summer. Wind also exhibits a diurnal pattern, with levels typically higher at night, reaching their peak in the early morning, and subsequently decreasing throughout the daylight hours. 

To focus on the temporal structure within one day and given the varying power production capacities of these wind farms, we perform a shift-and-stretch transformation as in \cite{ludkovski2022} relative to a target date (e.g., April $1^{st}$), taking care of the annual seasonality and varying data scales. Furthermore, we divide by asset capacity to obtain the forecasted $\Tilde{p}^F_{m,j,n}$ and actual $\Tilde{p}^A_{m,j,n}$ power \emph{ratios} scaled to the range $(0,1)$; with $m$, $j$, and $n$ denoting the index of location, hour, and day, respectively.

\subsection{Forecast Error}
\begin{figure}[!htb]
    \centering
    \begin{tabular}{rc} {\Large a)} & 
        \raisebox{-.25\height}{\includegraphics[width=0.85\textwidth]{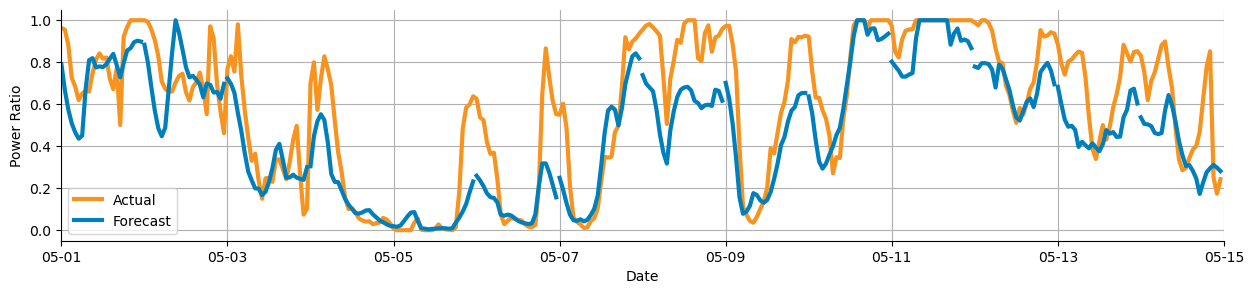}} \\
     {\Large b)}  & 
        \raisebox{-.25\height}{\includegraphics[width=0.85\textwidth]{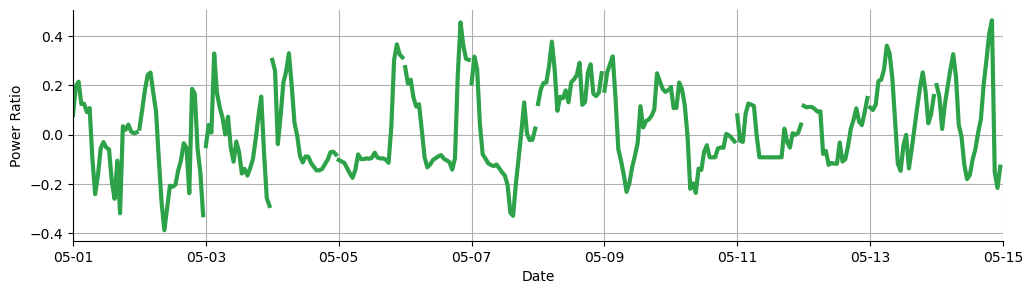}}
    \end{tabular}
    \caption{Power ratios at Ajax Wind Farm for every hour in May 2018 (336 hours total). 
    a): Actual $\Tilde{p}^{A}$ and forecasted $\Tilde{p}^{F}$ wind power ratios. b) Forecast errors $y$. $\Tilde{p}^A_{\cdot}$ is systematically higher than $\Tilde{p}^F_{\cdot}$ during this period. The forecast error $y_{\cdot}$ is symmetrically distributed around zero.}
    \label{fig: ajax_fa}
\end{figure}

We focus on modeling the forecast errors $y_{m,j,n}$  defined below in Equation \eqref{eq: fcst_error}.
We first compute the difference between the forecasted and actual power ratios:
\begin{equation}
   \Tilde{y}_{m,j,n}=\Tilde{p}^A_{m,j,n} - \Tilde{p}^F_{m,j,n}
\end{equation}
and then center the $\Tilde{y}$'s to ensure their mean is zero.
In general, forecasts are biased. For example, on a calm day the wind forecast is essentially zero, while the actual production is non-negative, so on average will be higher than the forecast, see Figure~\ref{fig: ajax_fa}a.  To center  $\Tilde{y}_{\cdot}$ we randomly divide the dataset days into a training set and a testing set.   The training set includes all 181 locations with 80\% of samples (24 hours across 87 days) while the testing set has 37 locations, 22 days. For each location $m$, we compute the mean, $\bar{{y}}_m$ over a period of 24 hours across 87 days in the training set. Then we get the forecast error as:
\begin{equation}
    y_{m,j,n} = \Tilde{y}_{m,j,n} - \bar{{y}}_m.
    \label{eq: fcst_error}
\end{equation}

Compared to the actual and forecast data, the forecast error rarely reaches the bounds and exhibits a symmetric distribution.
The typical standard deviation of forecast errors ranges between 0.135 and 0.268, with an average value of approximately 0.188. The distribution of forecast errors tends to have a slight right skew on average, with about 87\% of the locations exhibiting positive skewness.
This results in a distribution that more closely approximates a Gaussian distribution, better aligning with the assumptions of GP than the original datasets. Figure~\ref{fig: ajax_fa}b presents the forecast errors for the selected wind farm over the same time period as depicted in Figure~\ref{fig: ajax_fa}a. 

\subsection{Exploratory Data Analysis}

\begin{figure}[ht]
  \centering
\includegraphics[width=0.5\textwidth]{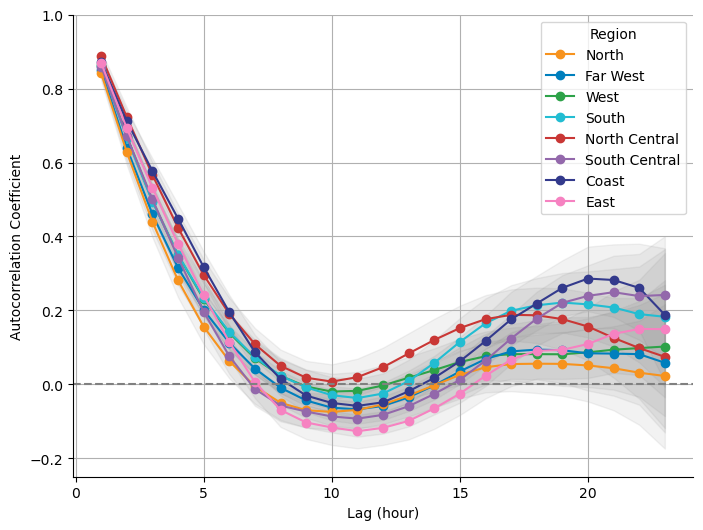}
  \caption{The autocorrelation of forecast errors at different hour lags within a single day. The regional ACF is calculated as the mean of ACFs of each location within that region. The grey region is the bootstrapped 95\% confidence interval of ACF.
  }
  
  \label{fig:acf_region}
\end{figure}

We next explore the  stylized features of the covariance structure in the resulting dataset of forecast errors $y$ across 181 distinct locations over 24 hours. As mentioned, different days are treated as i.i.d.~samples of the space-time field $y_{m,j,\cdot}$.
To determine whether we can model the covariance structure within the spatial and temporal dimensions separately, we assess if variations in the temporal domain are independent of spatial variations.  This is accomplished by constructing autocorrelation plots across different time lags for each spatial location. Our findings, depicted in Figure~\ref{fig:acf_region}, show a consistent pattern across all locations, with autocorrelation rapidly decreasing for lags less than 8-10 hours, and then increasing for lags over 12 hours. The consistent pattern across regions suggests a limited interdependence, supporting our use of a separable structure for the spatial and temporal variations. 


\begin{figure}[!htb]
  \centering
  \begin{subfigure}[b]{0.48\textwidth}
        \centering
  \includegraphics[width=\textwidth]{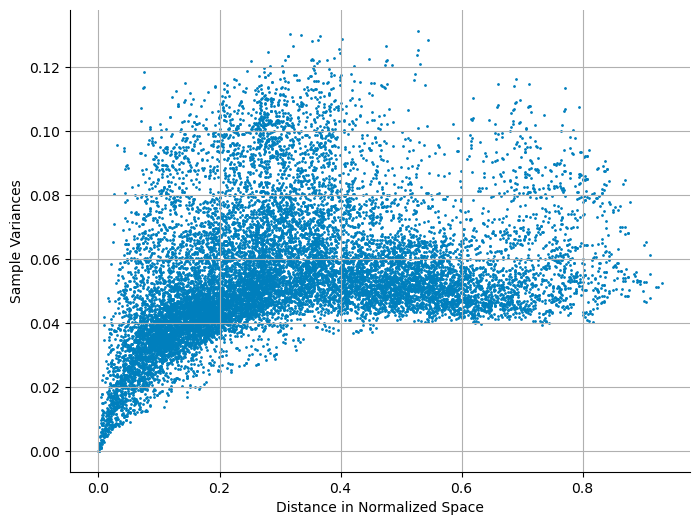}
  \caption{Spatial variogram, with distances calculated based on the $L_2$ norm in the normalized scale.}
  \label{fig: data_space_variogram}
  \end{subfigure}
  \hfill
  \begin{subfigure}[b]{0.48\textwidth}
        \centering
  \includegraphics[width=\textwidth]{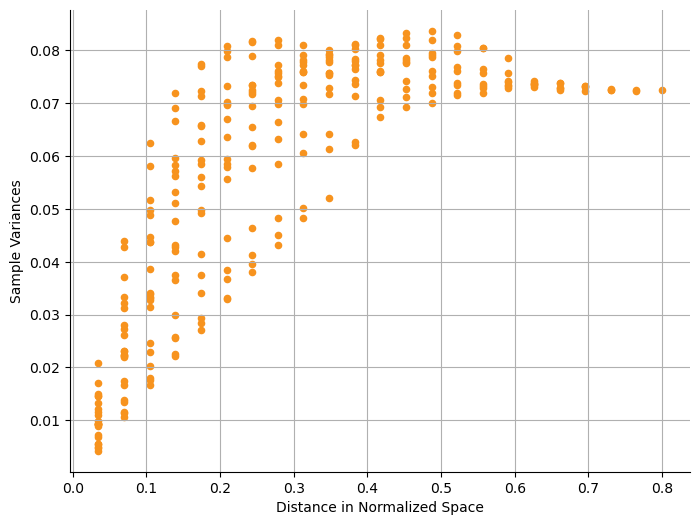}
  \caption{Temporal variogram, with distances calculated based on the $L_1$ norm in the normalized scale.}
  \label{fig: data_time_variogram}
  \end{subfigure}
  \caption{Variograms of the forecast error data. The pairs $(d, v)$ are visualized with the x-axis representing the distance and the y-axis representing the estimated variance.}
\end{figure}

Naturally, locations that are closer together tend to experience more similar weather and wind and hence have more correlated $y$'s. This suggests that spatial covariance of $y_{\cdot,t}$ is linked to the respective spatial distance. A graphical representation of the spatial dependence of the data is provided by the variogram \cite{cressie2015statistics}. For a dataset with a stationary spatial covariance structure, the variogram initially increases, indicating increasing dissimilarity with increasing distance between points, and then stabilizes asymptotically as the distance grows. In the real world, pairs of locations that share the same spatial distance may not necessarily exhibit the same level of similarity due to variations in the natural landscape. For example, the correlation between data from two wind farms built on mountains is likely to differ from that between two wind farms built on plains. Figure~\ref{fig: data_space_variogram}, visualizes the variogram of our forecast error data versus spatial distances. For two locations $\mathbf{s}_1$ and $\mathbf{s}_2$ in the normalized spatial domain, we construct the $(d, v)$ pair by first taking the $L_2$ norm between them, $d=\|\mathbf{s}_1-\mathbf{s}_2 \|_2$; then we compute the sample variance by taking the mean squared difference $v=\frac{1}{NT} \sum_{n=1}^N\sum_{t=1}^{T} (y_{1, t, n} - y_{2,t, n})^2$, across days $n=1, \ldots,N$ and hours $t=1, \ldots,T$. It is evident that the spatial variation does not exhibit a consistent pattern, since for a given spatial distance we observe widely different variogram values. This suggests that the data variation within the space dimension is not solely determined by the spatial distance between two locations. This observation highlights a significant spatial nonstationarity within our dataset, thus representing a primary challenge in our modeling analysis. 

Switching to the temporal dimension, the temporal variogram is shown in Figure~\ref{fig: data_time_variogram}. For two time points $t_1$ and $t_2$ in the normalized temporal domain, we construct $(d, v)$ pair by taking the $L_1$ norm between them, $d=\|t_1-t_2 \|_1$ and then computing the sample variance by taking the mean squared difference $v=\frac{1}{NM} \sum_{n=1}^N\sum_{m=1}^{M} (y_{m, t_1, n} - y_{m, t_2, n})^2$, across days $n=1, \ldots,N$ and locations $m=1, \ldots,M$. Two distinct patterns of data variation are clearly visible when the temporal distance is within the range $[0.2, 0.4]$. This indicates that the temporal covariance structure cannot be adequately described solely by time difference.

To recap, we validate using a separable covariance structure across time and space components but must address nonstationarity in both dimensions, as well as a periodic covariance pattern in time. We also emphasize that 
%
the spatial input dimension is continuous, $\mathbf{s} \in \mathbb{R}^2$, while the temporal dimension is gridded with each unit representing an hour, $t\in \{0:00, 1:00, ..., 23:00\}$.
The next section lays out our spatiotemporal GP model capturing this features with the ultimate formulation and discussion in Section~\ref{sec: model_overview}.

\section{Modeling}
\label{sec: modeling}

Let $\mathcal{X}$ represent our input domain, and $\mathbf{x}=(\mathbf{s}, t)$ denote an input point. To mitigate disparities in data scales across input domains, space and time, all dimensions of $\mathbf{x}$ are normalized to the range $(0,1)$ without reaching the bounds. This normalization facilitates the warping effect at the boundaries, which will be discussed in Section~\ref{sec: warp}. Notably, the spatial domain includes two coordinates: longitude and latitude. The normalization process is conducted while preserving the proportional relationship of spatial distances between locations. Consequently, pairs of locations equidistant in physical space maintain an equivalent distance in the normalized spatial domain.

Let $Y(\mathbf{s}, t)$ be the random variable denoting the forecast error 
where $\mathbf{s} \in \mathbb{R}^2$ is a pair of spatial coordinates, and $t\in \mathbb{Z}$ denotes the hour when the forecast error was measured. We assume the following structural equation model:
\begin{equation}
    Y(\mathbf{s},t) = f(\mathbf{s},t) + \bm{\epsilon}.
\label{eq:model}
\end{equation}
We model $f(\mathbf{s},t)$ as a GP, and $\bm{\epsilon}$ as a random variable with a Gaussian distribution that is independent of $\mathbf{s},t$ to account for noise or measurement errors in the observed data.  
Recall that a stochastic process $\{f(\mathbf{x}): \mathbf{x}\in\mathcal{X} \}$ is said to be a Gaussian Process, denoted as $GP(\mu(\mathbf{x}), k(\cdot, \cdot))$ if any finite set of $f(\cdot)$ values follow a multivariate Gaussian distribution. The marginal distribution is $Y(\mathbf{x}_i)\sim {\cal N}(\mu(\mathbf{x}_i), k(\mathbf{x}_i,\mathbf{x}_i)),$ and the joint is determined by the covariance function
$$\mathbb{E}[(Y(\mathbf{x}_i)-\mu(\mathbf{x}_i))(Y(\mathbf{x}_j)-\mu(\mathbf{x}_j))]= k(\mathbf{x}_j,\mathbf{x}_j),$$
that captures the dependencies between different inputs $\mathbf{x}_i, \mathbf{x}_j \in \mathcal{X}$. The GP kernel defines the shape and smoothness of $f$ and specifies the similarity between  inputs $\mathbf{x}_i$ and $\mathbf{x}_j$ in the input space \cite{genton2001classes}.

Constructing a conventional spatiotemporal GP model  involves integrating all time and location factors to establish the covariance matrix, resulting in a computational burden of $\cO(M^3\times T^3\times N^3)$ in our case, where $M$ is the number of spatial locations, and $T$ denotes the number of time points across $N$ days. Given the deseasonalization of the dataset, we posit independence among data points on different days, 
 constraining the temporal domain to the 24-hour duration of a single day. Consequently, we build a GP with only $\cO(M^3\times T^3)$ computational complexity:
\begin{equation}
    \mathbf{Y}_n \overset{\mathrm{i.i.d.}}{\sim} {\cal N}_{M\times T} (\mathbf{0}, \mathbf{K} + \sigma^2 \cdot \mathbf{I}),
\label{eq:distribution}
\end{equation}
where $\mathbf{Y}_n=(Y_{1,1,n}, Y_{1,2,n},..., Y_{1,T,n}, ...., Y_{M,T,n})^\top$, $n=1,2,...,N$, and $\sigma^2$ is the nugget hyperparameter, representing the variance of the observation noise. The prior mean function $\mu(\cdot)$ is set to be zero since we centered the forecast error by each location to have mean zero in the data preprocessing step. The covariance matrix $\mathbf{K}$ is decided by the structure of kernel $k(\cdot, \cdot)$ and GP hyperparameters $\bm{\theta}$. Section~\ref{sec: gp} reviews spatiotemporal GP and kernels. The statistical inference for our spatiotemporal GP is provided in Section~\ref{sec: mod_inference}. In Section~\ref{sec: warp} we review and summarize the input warping technique and its applications in our analysis.

\subsection{Spatiotemporal Gaussian Process Model}
\label{sec: gp}

In our analysis, the spatial input is continuous, while the temporal dimension is structured as a 24-hour grid, indicating varying degrees of smoothness 
in both time and space. Therefore, employing a separable kernel is a natural choice to accommodate these distinct characteristics. A separable kernel assumes that the covariance between two input domains can be decomposed as a product, with each component affecting the data independently,
$$k(\mathbf{x}_i,\mathbf{x}_j)=k_{\text{spatial}}(\mathbf{s}_i,\mathbf{s}_j) \cdot k_{\text{temporal}}(t_i,t_j)$$ where the  spatial kernel $k_{\text{spatial}}$ captures the spatial dependencies and interactions in the data, while the temporal kernel $k_{\text{temporal}}$ captures the temporal dependencies.

A stationary kernel assumes that the correlation between inputs depends only on their relative difference and can be written as a function of $\mathbf{x}_i-\mathbf{x}_j$, i.e., $k(\mathbf{x}_i,\mathbf{x}_j)=k_{\text{Stat}}(\mathbf{x}_i-\mathbf{x}_j)$, thus being translation invariant in the input space. Furthermore, when a stationary kernel depends only on the norm of the distance between the two inputs, it is said to be isotropic/homogeneous, and is thus only a function of $d = || \mathbf{x}_i-\mathbf{x}_j||$, i.e., $k(\mathbf{x}_i,\mathbf{x}_j)=k_{\text{Iso}}(||\mathbf{x}_i- \mathbf{x}_j||)$. The most common isotropic kernel is the Mat\'ern family, which is characterized by three parameters: process variance $\eta$, range $\rho$, and smoothness $\nu$. $\eta$ represents the variance of the GP and controls the overall scale of the variation in the data; $\rho$ determines the spatial and temporal extent over which the covariance between points is significant; $\nu$ controls the smoothness of the GP. As $\nu$ approaches infinity, the Mat\'ern kernel converges to the Squared Exponential (SE) kernel, which represents a completely smooth process. Smaller values of $\nu$ lead to more rough or wiggly functions. Common Mat\'ern kernels include:
\begin{itemize}
\label{sec: matern_ker}
    \item If $\nu=0.5$, $k_{M12}(d)= \eta\exp(-\frac{d}{\rho})$ (Mat\'ern 1/2);
    \item If $\nu=1.5$, $k_{M32}(d)= \eta(1+\frac{\sqrt{3}d}{\rho}) \exp(-\frac{\sqrt{3}d}{\rho})$ (Mat\'ern 3/2);
    \item If $\nu=2.5$, $k_{M52}(d)=\eta(1+\frac{\sqrt{5}d}{\rho}+\frac{5d^2}{3\rho^2}) \exp(-\frac{\sqrt{5}d}{\rho})$ (Mat\'ern 5/2);
    \item If $\nu\rightarrow \infty$, $k_{SE}(d)=\eta \exp(-\frac{d^2}{2 \rho^2})$ (Squared Exponential).
\end{itemize}

Presuming an isotropic structure enables more concise parameterizations of the kernel function. However, it may not adequately capture the variation patterns in real-world spatial and temporal data. To address this limitation, we propose a solution involving the mapping of the original input space to a latent space, where stationarity is maintained. This is achieved through the application of input warping on both space and time, as discussed in Section~\ref{sec: warp}.

Selecting an appropriate kernel function and tuning its hyperparameters are key steps to build the GP model that captures data patterns, as well as make accurate predictions or inferences. In Section~\ref{sec: kernel_eval} we employ synthetic examples to assess the efficacy of different kernel choices in our setting.

\subsection{Model Training and Prediction}
\label{sec: mod_inference}

The training process of the GP model involves hyperparameter estimation within the kernel and warping functions. We train the model by maximizing the log-marginal likelihood of the observed data given our model and its hyperparameters. This process utilizes optimization algorithms to iteratively adjust the hyperparameters, ensuring the best fit of the model to the data.

Considering our model as specified in Equations \eqref{eq:model} and \eqref{eq:distribution} with hyperparameters $\bm{\theta}$, the likelihood function of observed data $\mathbf{y}$ across $M$ different spatial locations and $T$ hours in $N$ days is: 
\begin{equation}
L(\bm{\theta}) = \prod_{i=1}^N p( \mathbf{y}_i  |\mathbf{0}, \mathbf{K} + \sigma^2 \cdot \mathbf{I}).
\end{equation}
Here, the covariance matrix $\mathbf{K}$ is defined by kernels $k(\cdot, \cdot)$ for both spatial and temporal inputs, with dimensions $M\times T$. Each daily sample is assumed to follow a multivariate normal distribution with probability density function $p(\mathbf{y}_\cdot)$. Since daily observations are i.i.d., we have a product of 
$N$ terms to construct the likelihood function.

To train a GP model based on observations $(\mathbf{X}, \mathbf{y})$, we  maximize the log-likelihood:
\begin{equation}\label{log-l}
\begin{split}
l(\bm{\theta}) &= -\frac{NMT}{2} \text{log} (2\pi) - \frac{N}{2} \text{log} |\mathbf{K} + \sigma^2 \mathbf{I}| -\frac{1}{2} \sum_{i=1}^N \mathbf{y}_i^\top (\mathbf{K} + \sigma^2 \mathbf{I})^{-1} \mathbf{y}_i
\end{split}
\end{equation}
to obtain Maximum Likelihood Estimates (MLE) $\hat{\bm{\theta}}$. We can then derive the predictive distribution for $Y^*$
at the input $\mathbf{x}^*=(\mathbf{s}^*, t^*)$:
\begin{equation}
    Y^* | \mathbf{x}^*, \mathbf{X}, \mathbf{y} \sim \mathcal{N}(\mu_*, \sigma_*^2 + \hat{\sigma}^2 )
\label{eq: log-l}
\end{equation}
Here, setting $\mathbf{k}_* = k(\mathbf{X},\mathbf{x}^*; \hat{\bm{\theta}})$, $k_{**} =
k(\mathbf{x}^*, \mathbf{x}^*; \hat{\bm{\theta}})$ the posterior (and also predictive) mean is:
\begin{equation}
    \mu_* = \mathbf{k}_*^\top(\mathbf{K} + \hat{\sigma}^2 \mathbf{I})^{-1}\mathbf{y},
\label{eq: pred_mean}
\end{equation}
and the posterior variance is: 
\begin{equation}
  \sigma_*^2 =k_{**} - \mathbf{k}_*^\top(\mathbf{K} + \hat{\sigma}^2  \mathbf{I})^{-1} \mathbf{k}_*.
  \label{eq: pred_cov}
\end{equation}

\subsection{Input Warping}
\label{sec: warp}

The input warping technique \cite{damian2001bayesian, sampson1992nonparametric, schmidt2003bayesian, snoek2014input} addresses output nonstationarity by transforming the original domain to a latent space where stationarity holds. Geographically, warping allows for the recognition and accommodation of spatial heterogeneity in wind patterns. Different regions with distinct geographical features, such as mountainous areas or coastlines, can be appropriately modeled, capturing the variability in wind behavior across diverse landscapes. Similarly, warping provides a more accurate representation of how wind conditions evolve throughout the day instead of treating time as a uniform dimension without considering variations in wind behavior over different hours.

Given a kernel $k(\cdot, \cdot)$ and warping function $g(\cdot)$, the covariance function between any two inputs $\mathbf{x}_i, \mathbf{x}_j \in \mathcal{X}$ is transformed to $k(g(\mathbf{x}_i), g(\mathbf{x}_j))$. The warping function $g(\cdot)$ has the form of a composition of $l \ge 1$ warping layers \cite{zammit2022deep, vu2023constructing}:
\begin{equation}
\begin{split}
    \label{eq: comp_warp}
    g(\mathbf{x}) 
    &= g_l \circ g_{l-1} \circ ... \circ g_1 (\mathbf{x}).
\end{split}
\end{equation}

The warping functions must be injective to facilitate hyperparameter estimation and enable the recovery of the original input locations. A frequently used class of warping functions is cumulative distribution function (CDF) based, e.g., Beta CDF \cite{snoek2014input} and Kumaraswamy CDF, projecting each input dimension of the input $\mathbf{x}$ through the CDF to a stationary latent space.

Another commonly used class of input warping functions involves the use of a combination of basis functions \cite{cressie2022basis}, which provides a general framework for modeling nonstationary processes. This class includes a collection of both linear and nonlinear basis elements, such as the axial warping unit \cite{zammit2022deep}. These functions facilitate adaptable transformations of inputs.
Among them, the radial basis function (RBF) unit \cite{buhmann2000radial} stands out for its flexibility, being suitable for warping inputs with one or more dimensions. It allows for deformation across multiple dimensions simultaneously and can expand or contract the input space locally around a center. Suppose the input $\mathbf{x}$ has $D$ input dimensions, the RBF $g_{\text{rbf}}(\cdot)$ is parameterized by weights $\mathbf{w}$, centers $\bm{\gamma}$, and scale $a$:
\begin{equation}
\label{eq: rbf}
    g_{\text{rbf}}(\mathbf{x}) =
    \begin{bmatrix}
     x_1 + w_1 \times(x_1 - \gamma_1)\text{exp}(-\frac{1}{2 a^2} \|\mathbf{x} - \bm{\gamma}\|^2 )\\
    \vdots \\
    x_D + w_D \times(x_D - \gamma_D)\text{exp}(-\frac{1}{2 a^2} \|\mathbf{x} - \bm{\gamma}\|^2 )
    \end{bmatrix}.
\end{equation}

\begin{figure}[htbp]
    \centering
    \begin{subfigure}[b]{0.31\textwidth}
        \centering
        \includegraphics[width=\textwidth]{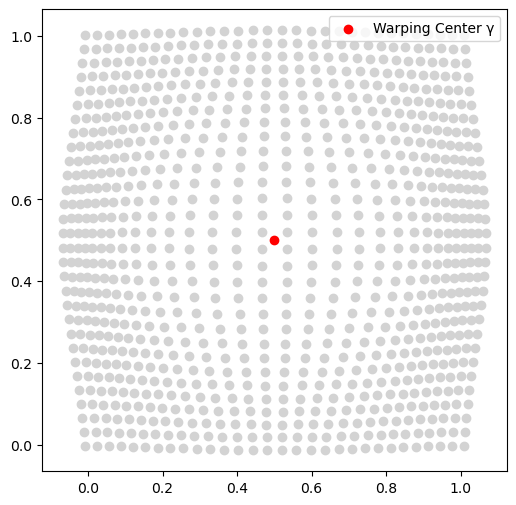}
        \caption{$a=0.25$, $\mathbf{w}=(1.0, 0.2)$,
        $\bm{\gamma}=(0.5,0.5)$.}
        \label{fig:subfig1}
        \begin{minipage}{.1cm}
            \vfill
            \end{minipage}
    \end{subfigure}
    \hfill
    \begin{subfigure}[b]{0.31\textwidth}
        \centering
        \includegraphics[width=\textwidth]{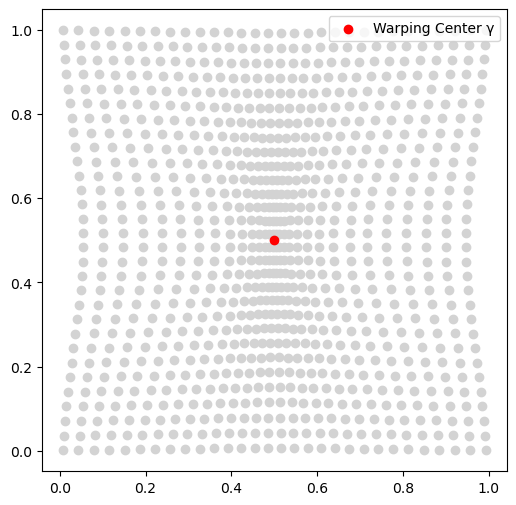}
        \caption{$a=0.25$,$\mathbf{w}=(-0.8, -0.1)$, $\bm{\gamma}=(0.5,0.5)$.}
        \label{fig:subfig2}
        \begin{minipage}{.1cm}
            \vfill
            \end{minipage}
    \end{subfigure}
    \hfill
    \begin{subfigure}[b]{0.31\textwidth}
        \centering
        \includegraphics[width=\textwidth]{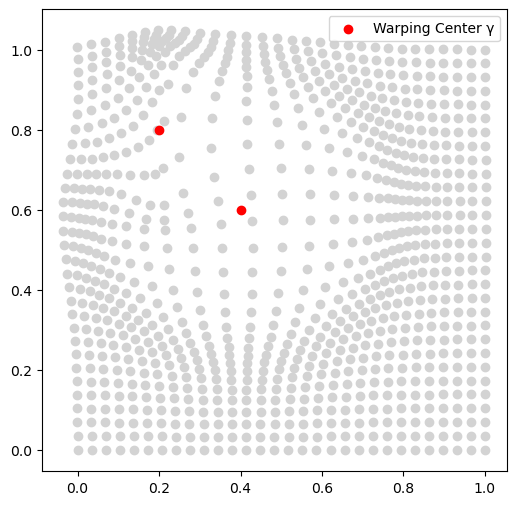}
        \caption{$a^{(1)}=0.18$, $\mathbf{w}^{(1)}=(1.2, 1.0)$, $\bm{\gamma}^{(1)}=(0.4,0.6)$; $a^{(2)}=0.10$, 
        $\mathbf{w}^{(2)}=(-0.7, 1.5)$, $\bm{\gamma}^{(2)}=(0.2,0.8)$.
        }
        \label{fig:two_rbf_eg}
    \end{subfigure}
    \caption{Illustration of input warping with RBF units $g_{\text{rbf}}$. We visualize how 
    a uniform $(0,1)\times(0,1)$ grid is transformed by one and two RBF warping units with indicated hyperparameters. }
    \label{fig:rbf_eg}
\end{figure}

The weights $w$ are limited within $(-1,\frac{1}{2}\exp(\frac{3}{2}))$ to ensure injectivity, a positive (resp.~negative) weight $w_d$ expands (resp.~contracts) dimension $d$.   The stretching transformation is carried out relative to the center $\bm{\gamma}$. The scale $a > 0$ governs the warping resolution, smaller values of $a$ allow for more localized deformation. Figure~\ref{fig:rbf_eg} illustrates examples of the warping effect induced by the RBF unit, as described by Equation \eqref{eq: comp_warp} in $D=2$ dimensions: Figures~\ref{fig:subfig1} and~\ref{fig:subfig2}  visualize the warping effect of one RBF unit, Figure~\ref{fig:two_rbf_eg} depicts warping with two compositional layers. Each layer independently transforms the uniform $(0,1)\times(0,1)$ grid centered around its respective location. Additionally, the two layers exhibit interaction with each other. The Figures visualize how a greater absolute value of the weight corresponds to a more pronounced deformation impact along the associated axis; similarly a larger value of scale $a$ shows a more global contraction (with negative weights $\mathbf{w}$) or expansion (with positive weights $\mathbf{w}$).

When constructing the warping structure, unlike Zammit et al. \cite{zammit2022deep} and Vu et al. \cite{vu2023constructing}, which fix all warping hyperparameters, we fit them as part of the model training process. 
To simplify inference, we restrict attention to warping via RBF units $g_{\text{rbf}}$. The number of warping layers is varied and fine-tuned manually. 
Thanks to working with a separable kernel, we may independently vary $l_S, l_T$ based on the patterns in each input domain. The  selection of the number of warping layers 
must also consider model complexity, as each layer introduces multiple parameters to be learned. In Section~\ref{sec: warp_sim}, we validate the above logic using  synthetic examples to show the impact of the warping function $g(\cdot)$ structure on model performance.

\subsection{Method Overview and Implementation}
\label{sec: model_overview}

We build a spatiotemporal GP model with inputs being 24 hours across 181 locations and a separable covariance:
\begin{equation}\label{eq:main-GP}
    \begin{split}
k((\mathbf{s}_i, t_i),(\mathbf{s}_j, t_j))&=\eta \times k_{\text{spatial}}(||g_S(\mathbf{s}_i) - g_S(\mathbf{s}_j)||_2) \times k_{\text{temporal}}(|t_i - t_j|, |g_T(t_i) - g_T(t_j)|),
    \end{split}
\end{equation}
where process variance $\eta$ controls the overall variation of the GP model, and $k_{\text{spatial}}$ and $k_{\text{temporal}}$ are isotropic kernels. Using a Mat\'ern family $k_{\text{spatial}}$,  $g_S(\cdot)$ is the spatial warping function, comprising one or more two-dim $g_{\text{rbf}}$ layers to address spatial nonstationarity. 


Since autocorrelation is closely connected to the covariance, the temporal kernel $k_{\text{temporal}}$ 
should emulate the pattern depicted in Figure~\ref{fig:acf_region}. 
To capture this shape, 
we adopt a Mat\'ern kernel augmented with a periodic component,
\begin{equation}
k_{\text{temporal}}(|t_i-t_j|, |g_T(t_i) - g_T(t_j)|)= k_M(|g_T(t_i) - g_T(t_j)| ; \rho_T) + \eta_p k_p(|t_i-t_j| ; \rho_p, p),
\label{eq: k_temp}
\end{equation}
with the periodic kernel defined as $$k_p(d; \rho_p, p)=\exp \Big(-\frac{2}{\rho_p^2}\sin^2(\frac{1}{2p} \pi d) \Big).$$ 
Finally, the temporal warping function $g_T(\cdot)$ is set to a single one-dim $g_{\text{rbf}}$ layer to capture the daily pattern.

To train our GP model, we maximize the log-likelihood in Equation \eqref{eq: log-l}.  Table~\ref{tab:params} summarizes the parameters of the GP kernel, warping, and nugget: $\bm{\theta} :=({\bm{\theta}}_k, {\bm{\theta}}_w, {\sigma}^2)$, specifying whether they are involved in the training step and outlining the constraints imposed during the optimization process.  We build isotropic spatial kernels from the Mat\'ern family, and a Mat\'ern plus periodic kernel \eqref{eq: k_temp} for temporal kernels. Therefore, we have 6 kernel parameters $\bm{\theta}_k=(\eta, \rho_S, \rho_T, \eta_p, \rho_p, p)$: $\eta$ signifies the overall variance of our spatiotemporal GP, while $(\rho_S, \rho_T)$ serve as range parameters representing smoothness characteristics of the spatial and temporal dimensions; parameters within the temporal kernel's periodic component $(\eta_p, \rho_p, p)$ are designed to account for the daily periodic pattern of the forecast error. 

As for warping, we have hyperparameters $\bm{\theta}_w=(\bm{\theta}_{w_S}, \bm{\theta}_{w_T})$. One spatial RBF warping layer includes 5 parameters to estimate, $\bm{\theta}_{w_S}=(\mathbf{w_S}=(w_1, w_2), \bm{\gamma_S}=(\gamma_1, \gamma_2), a_S)$. The temporal RBF warping layer includes 3 parameters to estimate, $\bm{\theta}_{w_T}=(w_T, \gamma_T, a_T)$. For example, a model with only temporal warping structure has 10 parameters to learn during the training process, whereas a model with additional one spatial warping layer has 15 parameters. To mitigate the issue of over-parameterization, we limit the number of temporal warping layers in the model to one, and -spatial warping layers to one or two in practice.

\begin{table}[ht]
\small
    \centering
    \begin{tabular}{c|cc}\toprule
        \textbf{Model Part} & \textbf{Parameter} & \textbf{Constraint} 
         \\
        \midrule
        \textbf{Kernel} 
        & Variance $\eta$ & $\eta > 0$ 
        \\
        \midrule
        \textbf{Mat\'ern}
            & Range $(\rho_T, \rho_S)$ & $\rho > 0$ 
            \\
        \midrule
        \multirow{3}{*}{\textbf{Periodic}} 
            & Variance $\eta_p$ & $\eta_p > 0$
            \\
            & Range $\rho_p$ & $\rho_p > 0$ \\
            & Period $p$ & 
            $p > 0$
            \\

        \midrule
        \multirow{3}{*}{\textbf{Warping}}
            & Weight $(\mathbf{w_S}, w_T)$ & $w \in (-1, \frac{\exp(1.5)}{2})$ 
            \\
            & Center $(\bm{\gamma}_S, \gamma_T)$ & $\gamma \in [0, 1]$ \\
            & Scale $(a_S,a_T)$ & $a > 0$ \\

        \midrule
        \textbf{Noise} 
        & Nugget $\sigma^2$
        & $\sigma^2 > 0$
        \\
        \bottomrule
    \end{tabular}
    \caption{Summary of hyperparameters and their constraints in model \eqref{eq:main-GP}. 
    }
    \label{tab:params}
\end{table}

For training the warping function, the weights $\mathbf{w}$ can be both positive (expansion) and negative (contraction). When $w_i = 0$, the warping effect degenerates, creating a zero-gradient effect in the likelihood. Consequently, gradient descent optimization becomes effectively trapped in a given orthant of $\mathbf{w}$: the sign of the estimated $\hat{w}$ is determined by the sign of the initial guess $\hat{w}^{(init)}$. To overcome this issue, we advocate running the Adam optimizer \cite{kingma2014adam} with multiple starting points, starting with different signs of the initial weights $\hat{w}^{(init)}$. Due to the differing lengths of positive and negative domains for $w$, the starting points are strategically positioned at the midpoint of each half-domain to allow for a balanced exploration of both positive and negative values. The global maximum of the log-likelihood is then determined by comparing across those multiple runs.

Our model implementation utilizes Python and the \texttt{PyTorch} framework.  The synthetic studies in Section~\ref{sec: synthetic} and real data analyses in Section~\ref{sec: data_analysis} were executed on a 2021 MacBook Pro, featuring the Apple M1 Pro chip and 16 GB of memory. The trained models are evaluated using metrics discussed in Section~\ref{sec: eval}. 


\section{Evaluation Metrics}
\label{sec: eval}

In this section, we summarize several evaluation metrics for model selection and predictive performance.
To evaluate the appropriateness of model choices, we use a modified Bayesian Information Criterion (BIC) to quantify both the model performance on the training dataset and the complexity of the model.
To evaluate model performance on the test set, we use the root mean squared error (RMSE) for point prediction and probabilistic forecast assessment methods introduced in Section~\ref{sec: pfa} for probabilistic prediction.
Let $\mathbf{y}$, $\mathbf{y}^*$ denote observations in the training and testing dataset,
$M$ and $T$ the number of  training spatial locations and training temporal hours and $M^*, T^*$ the number of test locations and test hours.

\subsection{Modified Bayesian Information Criterion}

We employ Bayesian Information Criterion (BIC) to score candidate models, with lower BIC being better. For a model with kernel $k$, BIC is set to:
\begin{equation}\label{eq:bic}
    \text{BIC}(k, \hat{\bm{\theta}}) = - l_k(\hat{\bm{\theta}}; \mathbf{y}) + \frac{|\hat{\bm{\theta}}_{w_S}|\text{log}(M)}{2} 
    + \frac{|\hat{\bm{\theta}}_{w_T}|\text{log}(T)}{2}
    + \frac{|\hat{\bm{\theta}}_k|\text{log}(MT)}{2},
\end{equation}
\noindent 
where $l_k(\bm{\theta}; \mathbf{y})=\text{log } p(\mathbf{y}| k, \bm{\theta})$ is the log-likelihood of  $\mathbf{y}$ under a given kernel $k$ calculated in Equation \eqref{log-l}, $\hat{\bm{\theta}}$ is the maximum likelihood estimate (MLE) of $l_k(\bm{\theta}; \mathbf{y})$,  $|\hat{\bm{\theta}}_{w_S}|$ (respectively $|\hat{\bm{\theta}}_{w_T}|$) is the number of estimated spatial (temporal) warping hyperparameters and $|\hat{\bm{\theta}}_k|$ is the number of trained kernel hyperparameters.  In a conventional BIC definition, the penalty term reflects the number of parameters in the model and the number of data points. In \eqref{eq:bic} we modify it to account for the fact that training days are i.i.d.~and hence do not impact model complexity. The penalty terms are thus: the first term penalizes the parameters associated with spatial warping and the number of spatial locations ($M$); the second one penalizes the parameters related to temporal warping and the number of hours ($T$); the third term penalizes the kernel parameters and the product $M T$, which represents the number of location-hour combinations and determines the size of the GP covariance matrix. 

\subsection{Probabilistic Forecast Assessment}
\label{sec: pfa}

A probabilistic forecast takes the form of a predictive probability distribution \cite{gneiting2014probabilistic,gneiting2023probabilistic}. Under our model, the predicted distribution for the input $(\mathbf{s}_m, h_t)$ on day $n$ is $Y_{m,t,n}\sim \mathcal{N}(\hat{\mu}_{{m, t, n}}, \hat{\sigma}^2_{m,t})$, where the model predicted variance $\hat{\sigma}^2_{m,t}$ for $\mathbf{x} = (\mathbf{s}_m, h_t)$ corresponds to the diagonal element in $\bm{\Sigma}^*+\hat{\sigma}^2 \mathbf{I}$. Notably, this predicted variance remains constant across different days for a fixed input $\mathbf{x}$, as indicated by Equation \eqref{eq: pred_cov}. To assess this predictive distribution, we apply several probabilistic metrics for uncertainty quantification and calibration.

\textbf{Root Mean Squared Error: }
A lower test RMSE indicates a better performance concerning the GP point prediction: 
\begin{equation}
\label{eq: RMSE}
    \text{RMSE} = \sqrt{\frac{1}{N^* M^* T^* } \sum_{m=1}^{M^*} \sum_{t=1}^{T^*}\sum_{n=1}^{N^*} (y^*_{m, t, n} - \hat{\mu}_{{m, t, n}})^2 },
\end{equation}
where the testing set has $N^*$ days and $\hat{\mu}_{{m, t, n}}$ is the model predicted mean for $\mathbf{x} = (\mathbf{s}_m, h_t)$ on day $d_n$, see Equation \eqref{eq: pred_mean}.

\textbf{PIT and KS Test: }
The probability integral transformation (PIT) \cite{dawid1984present, diebold1997evaluating} is used to assess the performance of a statistical model in terms of its predictive quantiles. 
Let \begin{equation}
\label{eq: pit}
q^*_{m,t, n}=\Phi \left(\frac{y^*_{m,t,n}-\hat{\mu}_{{m,t, n}}}{\hat{\sigma}_{m,t}} \right), 
\end{equation}
where $\Phi(\cdot)$ is the standard normal cumulative distribution function. We perform PIT to obtain  $\mathbf{q}^*$ across test locations, hours, and days and then compare $\mathbf{q}^*$ to the Uniform distribution by performing the Kolmogorov-Smirnov (KS) test \cite{berger2014kolmogorov}. Lower test statistics and larger p-value indicate better performance.


\textbf{Coverage: }
We use coverage \cite{dodge2003oxford} to evaluate the performance of predictive confidence interval estimation. For a test observation $y^*_{m,t,n}$  coverage is given by:
\begin{equation}
    C_{\alpha}(m,t,n) = \mathbf{1}_{\{\frac{\alpha}{2}\leq q^*_{m,t, n}\leq 1- \frac{\alpha}{2}\}}.
\end{equation}
The overall coverage statistic is:
\begin{equation}
    \label{eq: coverage}
    C_{\alpha}=\frac{1}{N^* M^* T^*}\sum_{m=1}^{M^*} \sum_{t=1}^{T^*}\sum_{n=1}^{N^*} C_{\alpha}(m,t,n).
\end{equation}
A model with a value of $C_{\alpha}$ closer to $100\times (1-\alpha) \%$ has better predictive performance.

\textbf{Average Interval Score: }
The interval score (IS) \cite{gneiting2007strictly} is a scoring rule to assess the performance of predictive intervals. For a level $\alpha \in [0,1]$, the respective $100\times (1-\alpha) \%$ predictive interval is $[l_{\alpha}, u_{\alpha}]= [\hat{\mu}_{{m, t, n}} - z_{\alpha/2} \hat{\sigma}_{m,t}, \hat{\mu}_{{m, t, n}} + z_{\alpha/2} \hat{\sigma}_{m,t}]$, where $z_{\alpha/2}$ is the $\alpha/2$ critical value for the standard normal distribution. For a test output $y$,
the interval score is
\begin{equation}
\begin{split}
    IS_{\alpha}(y)= (u_{\alpha}-l_{\alpha}) &+ \frac{2}{\alpha} (l_{\alpha}-y) \mathbf{1}\{y < l_{\alpha}\}\\
    &+ \frac{2}{\alpha} (y-u_{\alpha}) \mathbf{1}\{y > u_{\alpha}\}.
\end{split}
\end{equation}
A lower average IS indicates a better performance for the predictive interval:
\begin{equation}
    \text{Avg IS}_{\alpha} = \frac{1}{N^* M^* T^*}\sum_{m=1}^{M^*} \sum_{t=1}^{T^*}\sum_{n=1}^{N^*} IS_{\alpha} (l_{m,t,n}, u_{m,t,n}; y^*_{m,t, n}).
\end{equation}

\section{Model Validation Case Studies}
\label{sec: synthetic}

To validate our modeling setup before proceeding to analysis of the ERCOT NREL wind power dataset, we present in this section two synthetic experiments. The first experiment considers the ability of our framework to recover the underlying covariance structure, in particular the kernel family and its hyperparameters. The second experiment focuses on the ability to detect and learn input warping. 
Our goals are to ascertain the efficacy of the warping technique in mitigating nonstationarity and to assess the potential for accurately recovering the form of the warping function. 

In both case studies, we generate data based on a known spatio-temporal structure, namely a GP with a given kernel. We then fit several variants of our framework and compare the resulting estimated GPs to the ground truth. 
Models are evaluated according to metrics discussed in Section~\ref{sec: eval}. For coverage tests, we use reference level of $80\%$; the interval score is calculated based on the $95$\% predictive interval.

\subsection{Learning the spatial covariance structure}
\label{sec: kernel_eval}

In the first synthetic example, we focus on comparing various spatial kernel choices. The primary objective is to understand to what extent it is possible to distinguish and identify the true spatial kernel based on a synthetic dataset that is similar to our test one. 
Specifically, we generate a synthetic dataset for 27 locations $\mathbf{s}_j$ under a separable, isotropic kernel setting, with the temporal kernel set to be Mat\'ern-3/2 (denoted as M32) and the spatial kernel $M$ one of Squared Exponential (SE),  Mat\'ern-5/2 (M52), Mat\'ern-3/2 (M32), Mat\'ern-1/2 (M12) families, i.e., $M\in \{\text{SE, M52, M32, M12}\}$, discussed in Section~\ref{sec: matern_ker}. The 27 locations are a subset of the West zone in ERCOT, with 25 used for training and 2 for testing. The ground-truth generative distribution is 
$$\mathbf{Y} \sim \mathcal{N}_{27\times 24} (\mathbf{0}, \mathbf{K}_{\text{true}} + \sigma^2 \cdot \mathbf{I}),$$
where $\mathbf{K}_{\text{true}}$ is determined by:
\begin{equation*}
    \begin{split}
k((\mathbf{s}_i, t_i),(\mathbf{s}_j, t_j))
    &= \eta \times k_{M}(||\mathbf{s}_i - \mathbf{s}_j||_2; \rho_S) \times k_{M32}(|t_i - t_j|; \rho_T),
    \end{split}
\end{equation*}
with $\sigma^2 = 0.05, \eta = 0.03$, $\rho_S = 1.0, \rho_T = 2.0$.
The respective 
hyperparameters are $\bm{\theta}=(\eta, \rho_T, \rho_S, \sigma^2)$. Comparing BIC values among models is equivalent to comparing the negative log-likelihood, which is the training loss.

We simulate 109 days (i.e.~independent samples) for these 27 locations using four different spatial kernel choices: $M_{\text{True}} \in \{\text{SE, M52, M32, M12}\}$. The training set consists of 87 days of all 27 locations and the test set is samples of the other 22 days of 2 locations. For each dataset, we then fit four models $M_{\text{Model}} \in \{ \text{SE, M52, M32, M12}\}$, each model utilizing the respective spatial kernel.  This gives a total of $4 \times 4$ models based on all combinations of $M_{\text{True}}$ and $M_{\text{Model}}$.

Table~\ref{tab: ker_eval_metric1} and~\ref{tab: ker_eval_params} in the Appendix shows the true and model training loss, as well as the estimated parameters including lengthscales $\hat{\bm{\rho}}=(\rho_T, \rho_S)$ and variance $\hat{\eta}$ across these 16 testbeds. Across all four simulated datasets, models employing the same kernel as the data-generating process, $M_{\text{True}} = M_{\text{Model}}$ outperform the other three, achieving the lowest training loss. Notably, the SE, M52, and M32 demonstrate comparable training losses when the true and model kernels are among these three. However, when the true spatial kernel is M12, models using the other three kernels exhibit significantly higher training loss values. All 16 trained models successfully recover the nugget term $\sigma^2$. 

Next, we evaluate 16 trained models on our test set, presented in Table~\ref{tab: ker_eval_metric1} and~\ref{tab: ker_eval_metric2} in the Appendix. The SE, M52, and M32 kernels present similar test RMSE values when both the true and model kernels are among these three. However, a model using the M12 kernel has a higher testing RMSE when the true kernel is among the other three. Furthermore, when the true kernel is M12, models employing the other three kernels display significantly higher RMSE values. In terms of probabilistic predictive performance, we confirm the consistent pattern that the M12 kernel exhibits a different performance pattern compared to the other three.

In this experiment, we constructed 16 models, each employing different kernels. Notably, the training loss shows sensitivity to diverse kernel choices. Models characterized by lower training loss values also manifest reduced test point predictive errors, making them preferable when precision in point prediction is a priority. The choice of kernel family shows little influence on predictive variance. When developing models for real-world data, we recommend the construction of at least two models: the first with kernels such as SE, M32, and M52, given their comparable performances on both training and testing sets; and the second with an M12 kernel, distinguished by a notably different pattern compared to the other three.

\subsection{Warping Function Recovery}
\label{sec: warp_sim}

In the second synthetic study, we investigate the effectiveness of input warping in addressing nonstationarity. The ground truth incorporates known spatial warping and we fit both warped and non-warped GPs. First we address whether a warped GP provides a better fit compared to a non-warped one. Second, we assess the relative performance of models as the number of warping layers is varied. 

The synthetic data is simulated for the same 27 locations in Section~\ref{sec: kernel_eval} in the West zone of ERCOT, maintaining the same setup except for adding spatial nonstationary through a known warping:
\begin{equation*}
    \begin{split}
k((\mathbf{s}_i, t_i),(\mathbf{s}_j, t_j))
    &= \eta \times k_{SE}(\|g(\mathbf{s}_i) - g(\mathbf{s}_j)\|_2, \rho_S=1.0) \times k_{M32}(|t_i - t_j|, \rho_T=2.0)
    \end{split}
\end{equation*}
\noindent where the spatial warping function $g(\cdot)$ has a compositional form of several RBF layers. Simulated data is generated directly on the standardized space-time domain and  features a separable kernel with a M32 temporal component and an SE spatial one. We generate wind generation ratios for 27 locations across 109 independently sampled days with 87 days (80\%) used for training and the rest for testing.

\textbf{Case Study $W_1$: Single Warping Layer.} In this case study, we aim to demonstrate that incorporating a warping structure effectively addresses data nonstationarity. Simulation data is generated by sampling from a GP with a single RBF warping unit (see Equation \eqref{eq: rbf}) for spatial locations only, $g(\mathbf{s})=g_{\text{rbf}}(\mathbf{s})$.

\begin{figure}[!htb]
\begin{subfigure}[b]{0.475\textwidth}
  \centering
\includegraphics[width=0.8\textwidth]{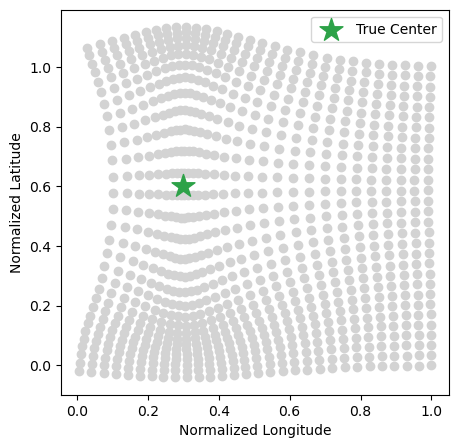}
  \caption{The true setting of warping of simulation data generation. The green star indicates the true position of the warping center $\bm{\gamma}$.}
  \label{fig: W1-true}
  \end{subfigure}
  \hfill
  \begin{subfigure}[b]{0.475\textwidth}
  \centering
\includegraphics[width=0.8\textwidth]{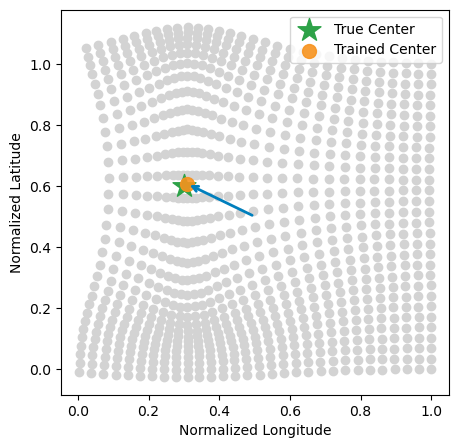}
  \caption{The blue arrow indicates the direction from the initial position of $\bm{\gamma}$ to its trained position, shown as a orange dot.}
  \label{fig: W1-model}
  \end{subfigure}
  \caption{Comparison of the estimated and true warping effects in the normalized spatial domain, viewed on a uniform $[0,1] \times [0,1]$ grid. 
  }
\end{figure}

 Table~\ref{tab: W2_train_tables} in the Appendix shows the ground truth hyperparameters vis-a-vis the model trained hyperparameters. We build two models: $W_1^{(M_1)}$ that includes one RBF layer and hence has  9 hyperparameters to train and $W_1^{(M_0)}$ with no warping, and four hyperparameters. Both of them successfully recover the nugget term. However, model $W_1^{(M_1)}$ provides a closer estimate of the variance parameter. It also successfully recovers the range parameter values and attains a similar BIC value as $W_1^{(M_0)}$, indicating a comparable training performance and thus does not introduce excessive training complexity.
 For the warping component, Model $W_1^{(M_1)}$ accurately estimates the hyperparameters in the warping function and successfully reconstructs the warped space, see Table~\ref{tab: W2_train_tables} and Figure~\ref{fig: W1-model}. 

\begin{table}[!htb]
    \centering
    \small
    \begin{tabular}{c | c c c  c}\toprule
    \textbf{Setting}
    & \textbf{RMSE}
    & \textbf{KS Test}
    & $\mathbf{C}_{0.2}$ 
    & $\textbf{Avg IS}_{0.05}$
    \\\midrule
         \multicolumn{5}{c}{Single Warping Layer Run} \\
    \midrule
        $W_1^{(M_1)}$
        & $\mathbf{0.0511}$  
        & $\mathbf{D=0.031, p=0.26}$
        & $78.41\%$
        & $\mathbf{0.2364}$
        
        \\
        $W_1^{(M_0)}$ 
        & $0.0516$
        & $D=0.034, p=0.16$
        & $\mathbf{78.88\%}$
        & $0.2392$
        
        \\  \midrule
         \multicolumn{5}{c}{Double Warping Layer Run} \\
       
          \midrule
        $W_2^{(M_2)}$
        & $\mathbf{0.051088}$ 
        & $\mathbf{D=0.030, p=0.31}$
        & $78.60\%$
        & $\mathbf{0.236091}$

        \\
        $W_2^{(M_1)}$
        & $0.051089$  
        & $D=0.031, p=0.27$
        & $78.50\%$
        & $0.236394$
        
        \\
        $W_2^{(M_0)}$ 
        & $0.051551$
        & $D=0.034, p=0.16$
        & $\mathbf{78.88\%}$
        & $0.239177$
        
        \\

        \bottomrule
    \end{tabular}
    \caption{Comparison of model performances on the test set. 
    \label{table: W1-W2-Test}}
\end{table}

The top half of Table~\ref{table: W1-W2-Test} summarizes model performances on the test set. 
The model with warping structure, $W_1^{(M_1)}$, shows enhanced accuracy in both point and probabilistic predictions,  evidenced by a lower test RMSE, lower average IS and a higher KS test p-value. In sum, incorporating the warping structure substantially improves both goodness-of-fit and testing performance and hence is effective in mitigating data covariance nonstationarity. 


\textbf{Case Study $W_2$: Double Warping Layers.}
In this case study, we aim to explore the effect of varying the number of warping layers on model performance in the presence of complex nonstationarity in the data. The synthetic data is generated using a two-layer RBF warping unit and no temporal warping. For a given location $\mathbf{s}$, the warping function is thus
$$g_S(\mathbf{s})=g_2\circ g_1 (\mathbf{s}),$$ with $g_1$ and $g_2$ as in \eqref{eq: rbf}; the respective warping parameters $(\mathbf{w}^1, \mathbf{w}^2, \bm{\gamma}^1, \bm{\gamma}^2, \mathbf{a})$ and three kernel hyperparameters $(\eta, \rho_T, \rho_S)$ are in Table~\ref{tab: W2_train_tables} in the Appendix. 

\begin{figure}[!htb]
\begin{subfigure}[b]{0.3\textwidth}
  \centering
\includegraphics[width=\textwidth]{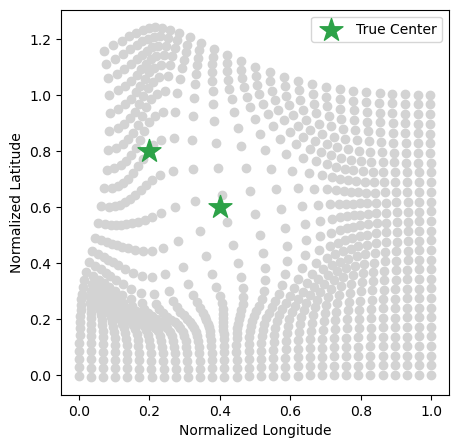}
  \caption{True warping of the synthetic dataset with two RBF units. The green stars show the ground truth warping centers $\bm{\gamma}$.}
  \label{fig: W2-true}
  \end{subfigure}
  \hfill
  \begin{subfigure}[b]{0.3\textwidth}
  \centering
\includegraphics[width=\textwidth]{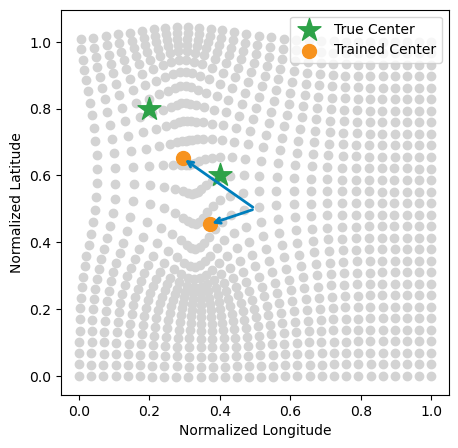}
  \caption{Warped space of model $W_2^{(M_2)}$. The orange dots indicate locations of the trained warping centers $\hat{\bm{\gamma}}_1, \hat{\bm{\gamma}}_2$.}
  \label{fig: W2-M2}
    \begin{minipage}{.5cm}
            \vfill
    \end{minipage}
  \end{subfigure}
  \hfill
  \begin{subfigure}[b]{0.3\textwidth}
  \centering
\includegraphics[width=\textwidth]{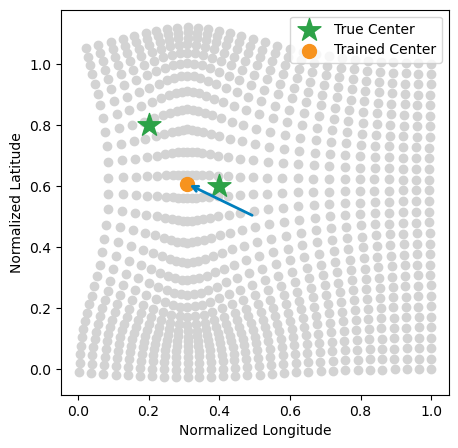}
  \caption{Warped space of model $W_2^{(M_1)}$. The orange dot shows the trained warping center $\hat{\bm{\gamma}}$.}
  \label{fig: W2-M1}
  \begin{minipage}{.1cm}
            \vfill
    \end{minipage}
  \end{subfigure}
  \caption{Visualizations of the fitted and true warping effects on the normalized spatial domain, shown in the form of a uniform $[0,1] \times [0,1]$ grid.}
  \label{fig: W2-model}
\end{figure}

We then train three GP models $W_2^{(M_i)}$, with $i=0,1,2$ RBF warping layers respectively. All three models utilize a separable kernel with M32 in the time dimension and SE in the space dimension. Consistent with the findings of the first study above, the model $W_2^{(M_0)}$ that lacks a warping structure mis-estimates the variance and range parameters relative to the ground truth. Although models incorporating warping, $W_2^{(M_2)}$ and $W_2^{(M_1)}$, do not perfectly recover the warped space (see Table~\ref{tab: W2_train_tables} and the visual comparison of the fitted warping effect in Figure~\ref{fig: W2-model}), both provide close estimates of the kernel hyperparameters. Between them, the one-layer warping model demonstrates better training performance, evidenced by a lower BIC value. 

Regarding performance on the test set, models with warping outperform the model without warping across most metrics, see bottom half of Table~\ref{table: W1-W2-Test}. Furthermore, the one-layer model $W_2^{(M_1)}$ exhibits comparable performance to the two-layer $W_2^{(M_2)}$ while maintaining reduced model complexity.

Based on the results presented, we confirm that although recovering the warping parameters is challenging, incorporating the warping structure in the model effectively addresses the issue of nonstationarity and enhances model performance. Our findings indicate that increasing the number of warping layers does not significantly improve model performance but introduces additional complexity. Consequently, for real data analysis, we recommend utilizing a warping structure with at least one RBF layer to address nonstationarity. However, the number of warping layers should be determined by the size of the dataset to avoid over-parameterization.

\section{Application to ERCOT Wind Energy Dataset}
\label{sec: data_analysis}

The analyzed  data set includes 109 consecutive days, from February $6^{th}$ to May $25^{th}$ 2018, of day-ahead forecasts and actual wind power ratios at 181 wind farm locations in Texas. Random allocation is employed to partition the dataset into training and testing subsets. 37 locations are designated for testing purposes. Moreover, the 109 days' samples for these 37 locations are randomly partitioned into training and testing sets, with 20\% allocated for testing and the remaining 80\% reserved for training purposes. To summarize, the training set comprises 87 days from all 181 locations, whereas the test set consists of 22 additional days across 37 testing locations. 

Augmented with a periodic term, the temporal kernel is set to be M32 warped by a one-dimensional RBF unit. The spatial kernel is set to SE or M12 based on the synthetic results discussed in Section~\ref{sec: kernel_eval}. To address spatial nonstationarity, we incorporate 1-3 two-dimensional RBF units into the spatial warping. In total, we build 10 models with different spatial kernel choices and warping structures, shown in Table~\ref{tab: data_metric_summary}. 

As a first finding,  we observe significant improvements on all predictive and training performance metrics for models with 
 both spatial and temporal warping (starting from SE-1-1). The results unambiguously show that both dimensions need to be warped, see for example dramatic improvements in BIC, RMSE and coverage metrics. Secondly, among the latter models,  
 %
we obtain consistent values for the estimated kernel parameters, with variances $\eta$ around 0.04-0.06, lengthscales $\rho_T$ around 0.1, and $\rho_S$ around 0.02 for the SE kernel and below 0.1 for the M12 kernel. This consistency suggests that similar stationarity is achieved in the warped spaces of these models. Although all models underestimate uncertainty (realized coverage below the nominal 80\%), several of the warped models get almost to nominal level with $C_{0.2} \in[ 0.70,0.75]$. Similarly, while none of the models pass the KS PIT test at $p=0.05$ level, the best ones achieve reasonable $D$ values. 

Thirdly, the M12-2-1 model with Mat\'ern-1/2 spatial kernel and two warping layers, and Mat\'ern-3/2 temporal kernel and single warping layer, demonstrates notably superior performance in terms of prediction accuracy and uncertainty.  Although the SE-1-1 model has the lowest RMSE (0.153) on the testing set, M12-2-1 exhibits a comparable prediction error (0.156). Moreover, M12-2-1 outperforms all other  models on probabilistic forecast assessment metrics. Relative to models with fewer warping terms, M12-2-1 achieves a lower BIC, indicating superior training performance without adding excessive model complexity. Conversely, it outperforms SE-3-1 and M12-3-1 models that add even more layers but end up with higher BIC, KS score and IS metric.

\begin{figure}[!htb]

\begin{minipage}[b]{0.55\textwidth}
\includegraphics[width=0.95\textwidth]{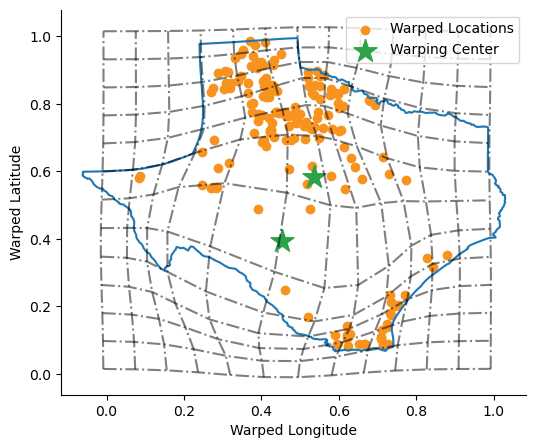}
\end{minipage}
  \begin{minipage}[b]{0.4\textwidth}
\includegraphics[width=0.95\textwidth]{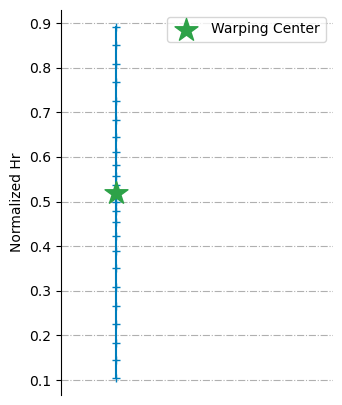}
\end{minipage}
  \caption{Visualization of warping effects in the proposed M12-2-1 model. Left: Spatial warping with 2 RBF layers; Right: Temporal warping with single RBF layer. Both plots use the normalized coordinates.}
  \label{fig: model_warp}
\end{figure}

Figure~\ref{fig: model_warp} visualizes the warping effects of the selected M12-2-1 model. The transformation on the spatial domain is performed around two locations in central and western Texas, expanding the southern part of Great Plains while condensing  the northern Great Plains, North Central Plains, as well as the southern Gulf Coast plains. Distances within the southern part of the Far West and West regions are stretched, while those within the South, North, North Central and rest of Far West and West regions are squeezed.  Figure~\ref{fig: model_scorr} in the Appendix illustrates the spatial correlation between two reference assets, Aguayo and Whirlwind, and other locations across Texas. The contour lines represent levels of correlation; due to the warping effects, different locations exhibit varying correlation patterns in their neighborhood. The correlation for the Aguayo wind farm drops off more sharply along the latitude, particularly to the south. We also observe compressed correlation to the west of Aguayo compared to its east. In contrast, Whirlwind is far from the fitted warping centers $\hat{\bm{\gamma}}_{1,2}$ and as a result the respective neighborhood correlation pattern is nearly elliptical, decreasing at a similar rate in all directions.

Temporally, warping yields a contraction centered around noon. Hence, mid-day hours are more correlated relative to morning or evening hours.  Figure~\ref{fig: model_tcorr} in the Appendix shows the corresponding temporal correlation of M12-2-1 relative to 4 AM. Moving away from the reference hour, the correlation first drops rapidly and then slowly decreases to zero because of the periodic effect. Decorrelation occurs after approximately 6 hours: the estimated temporal lengthscale $\rho_T=0.10$  corresponds to about 2.5 hours; for M32 kernel decorrelation is achieved roughly beyond two lengthscales.

\begin{table}[!htb]
    \centering
    \small
    \begin{tabular}{r c | c c c c c c c c}\toprule
    \textbf{Model}
    & $n_{hyp}$
    & \textbf{BIC}
    & \textbf{$\eta$}
    & \textbf{$(\rho_T, \rho_S)$} 
    & \textbf{$\sigma$}
    & \textbf{RMSE} 
    & \textbf{KS Test}
    & $\mathbf{C}_{0.2}$
    & $\textbf{IS}_{0.05
    }$
    \\
    \midrule
        SE-0-0 
        & $7$
        & $-3475.08$ 
         & $0.105$ 
        & $(0.11, 0.10)$
        & $0.0950$
        & $0.427$
        & $D=0.269$
        & $30.70$\%
        & $6.40$

        \\
        M12-0-0 
        & $7$
        & $-4343.34$ 
        & $0.100$ 
        & $(0.06, 0.79)$
        & $0.0267$
        & $0.395$
        & $D=0.340$
        & $25.99$\%
        & $6.51$

        \\
        SE-0-1 
        & $10$
        & $-3914.43$ 
         & $0.036$ 
        & $(0.10, 0.07)$
        & $0.0719$
        & $0.284$
        & $D=0.261$
        & $32.96$\%
        & $3.98$

        \\
        M12-0-1 
        & $10$
        & $-4111.89$ 
        & $0.116$ 
        & $(0.11, 0.98)$
        & $0.0841$
        & $0.373$
        & $D=0.265$
        & $35.62$\%
        & $5.04$
    
        \\
        SE-1-1 
        & $15$
        & $-5639.32$ 
         & $0.058$
        & $(0.07, 0.02)$
        & $0.0017$
        & $\mathbf{0.153}$
        & $D=0.074$
        & $70.94$\%
        & $1.30$

        \\
       M12-1-1 
        & $15$
        & $-5596.73$ 
        & $0.044$ 
        & $(0.08, 0.08)$
        & $0.0107$
        & $0.168$
        & $D=0.078$
        & $65.20$\%
        & $1.23$

        
         \\
        SE-2-1 
        & $20$
        & $-5316.90$ 
         & $0.062$ 
        & $(0.13, 0.02)$
        & $0.0618$
        & $0.160$
        & $D=0.073$
        & $69.58$\%
        & $1.36$

        \\
        \textbf{M12-2-1} 
        & $20$
        & $\mathbf{-5675.14}$ 
        & $0.052$ 
        & $(0.10, 0.05)$
        & $0.0021$
        & $0.156$
        & $\mathbf{D=0.041}$
        & $\mathbf{74.05}$\%
        & $\mathbf{1.01}$
        
        \\
        SE-3-1 
        & $25$
        & $-5627.56$ 
         & $0.056$ 
        & $(0.13, 0.01)$
        & $0.0061$
        & $0.158$
        & $D=0.069$
        & $73.03$\%
        & $1.37$

        \\
        M12-3-1 
        & $25$
        & $-5634.96$ 
        & $0.038$ 
        & $(0.13, 0.08)$
        & $0.0322$
        & $0.193$
        & $D=0.116$
        & $58.26$\%
        & $1.59$
        
        \\

        \bottomrule
    \end{tabular}
    \caption{Model performance for the ERCOT wind dataset across different kernels and warping choices. The first column presents the model setting. For example, in SE-0-1, the spatial warping kernel is SE, with no spatial warping layer and one temporal warping layer.}
    \label{tab: data_metric_summary}
\end{table}

To validate the consistency of model performance, we adopt a ``cross-validation" method, randomly dividing the 181 locations into leave-one-out folds. Specifically, we  train 5  models that use 4 out of the 5 folds for training and the fifth one as the test set. The five train-test splits differ from those for Table~\ref{tab: data_metric_summary} to ensure the replicability of our results under the same model setting. All of them follow the M12-2-1 model structure: an M12 spatial kernel and M32 plus periodic temporal kernel, with one RBF warping layer in the temporal domain and two layers in the spatial domain. Note that comparing BIC values across folds is equivalent to comparing their training loss. 

\begin{table}[!htb]
    \centering
    \small
    \begin{tabular}{c|c c c c c c c c}\toprule
    \textbf{Fold}
    & \textbf{Loss}
    & \textbf{$\eta$}
    & $(\rho_T, \rho_S)$ 
    & \textbf{$\sigma$}
    & \textbf{RMSE} 
    & $\mathbf{C}_{0.20}$
    & \textbf{KS Test}
    & $\textbf{IS}_{0.05
    }$
    \\
    \midrule
        1
        & $-5711.66$
        & $0.051$ 
        & $(0.10, 0.05)$
        & $0.0032$ 
        & $0.160$
        & $72.6\%$
        & $D=0.043$
        & $1.05$

        \\
        2 
        & $-5831.15$
        & $0.046$ 
        & $(0.13, 0.06)$
        & $0.0195$
        & $0.193$
        & $62.9\%$
        & $D=0.105$
        & $1.42$
    
        \\
        3
        & $-5782.90$
        & $0.040$ 
        & $(0.09, 0.07)$
        & $0.0010$
        & $0.171$
        & $64.7\%$
        & $D=0.082$
        & $1.32$

        \\
        4
        & $-5787.95$
        & $0.051$ 
        & $(0.10, 0.06)$
        & $0.0010$
        & $0.154$
        & $72.4\%$
        & $D=0.077$
        & $1.04$
        
        \\
        5 
        & $-5745.83$
        & $0.052$ 
        & $(0.10, 0.06)$
        & $0.0033$
        & $0.181$
        & $67.6\%$
        & $D=0.064$
        & $1.22$
        \\

        \bottomrule
    \end{tabular}
    \caption{Cross-validated training and testing results. Each folded model is trained using 80\% of the locations and uses the rest 20\% as the testing set. }
    \label{tab: data_cv}
\end{table}

Table \ref{tab: data_cv} presents the performance across the five folds. 
The model variances $\eta$ are in the range $[0.040, 0.52]$; lengthscales $\rho_T$ around 0.1 and $\rho_S$ around $0.06 \pm 0.01$, consistent with the M12-2-1 row in Table~\ref{tab: data_metric_summary}. Hence, the recovered GP hyperparameters are meaningful and stable as the training set is varied. The performance metrics are also consistent across the different train-test splits, with fold \#2 being somewhat of an outlier (cf.~its higher $\sigma$ estimate and worse coverage, KS statistic and interval score).  

\subsection{Model Based Simulations}

The developed framework is designed to simulate wind generation for day-ahead planning purposes. To showcase, 
we generate scenarios based on the the fitted M12-2-1 model with warping described above for various dates from 02/06/2018 to 05/25/2018. The scenarios are generated jointly across multiple assets and represent counterfactual realizations of wind power ratios on those days, conditioned on the given exogenous forecast. Starting with a single-asset perspective, we first illustrate such scenarios when conditioning both on the location-specific day-ahead forecast as well as realized generation from assets in the ``training'' set. Namely, considering 
asset $\mathbf{s}(m^*)$ and date $d^*$ within the test set, we sample 1000 scenarios from the GP predictive distribution conditional on observed data, i.e., realized forecast errors $y_{m,t,d^*}$ on date $d^*$ of $m=1, \ldots, 144$ assets that are not in the test set (see Equation~\eqref{eq: pred_mean} and~\eqref{eq: pred_cov}). By using the actual production ratios from neighboring locations, we can quantify the uncertainty of assets that are not directly observed.

\begin{figure}[!htb]
\begin{minipage}[b]{0.326\textwidth}
\includegraphics[width=0.95\textwidth]{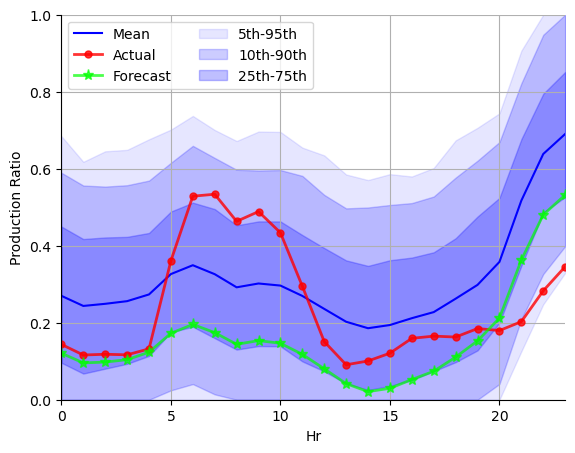} \\
\hspace*{35pt} Aguayo 02/06
  \end{minipage}
 \begin{minipage}[b]{0.326\textwidth}
\includegraphics[width=0.95\textwidth]{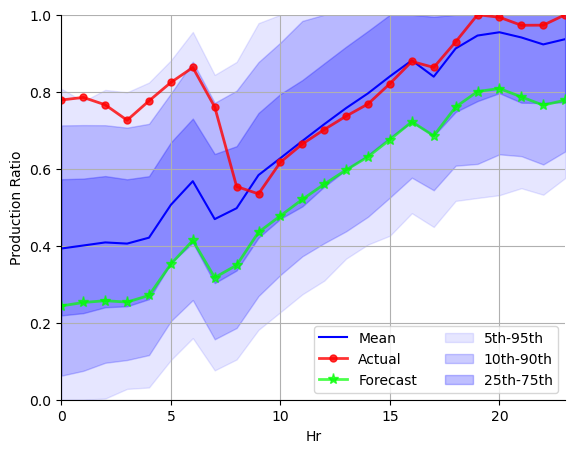} \\
\hspace*{35pt} Aguayo 04/11
  \end{minipage}
  \begin{minipage}[b]{0.326\textwidth}
\includegraphics[width=0.95\textwidth]{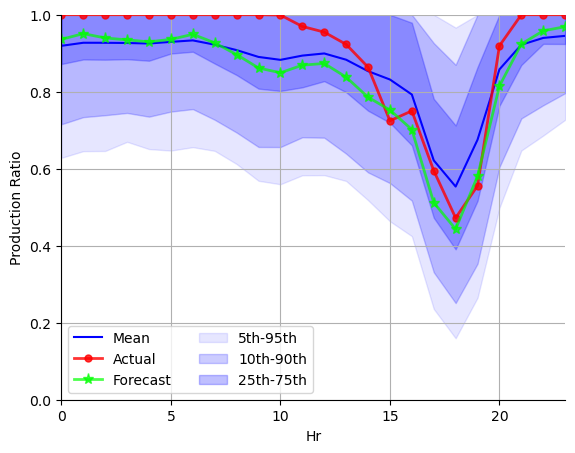} \\
\hspace*{35pt} Whirlwind 04/11
  \end{minipage}
  \caption{Representative asset-level scenarios of wind power ratio simulated from the fitted M12-2-1 model. We show 2 assets on 2 different days.
  \label{fig: scenario_asset}}
\end{figure}

Figure~\ref{fig: scenario_asset} shows the hourly-based single-asset simulations for Aguayo and Whirlwind assets, located in the North Central and North regions respectively, on a couple of representative days in the testing set. The figure shows the original forecast $\Tilde{p}^{F}_{m^*,\cdot,d^*}$, the actual wind power ratio $\Tilde{p}^{A}_{s^*,\cdot, d^*}$, and the mean of GP-based simulations. We also plot the model quantile bands at levels $\alpha=0.10, 0.20, 0.50$,  calculated by sorting the hourly simulated values and then saving the respective $\frac{1-\alpha}{2} \cdot 100\%$ and $\frac{1+\alpha}{2} \cdot 100\%$ quantiles.  Since the GP model generates forecast errors relative to $\Tilde{p}^F$, we clip the resulting $Y_{m^*,t,d^*} + \Tilde{p}^F_t + \bar{y}_{m^*}$ to ensure it falls within $[0,1]$.  We observe that the 10-90\% quantile band is roughly 0.6 in width, i.e.,~the production ratio is often $\pm 30\%$ above or below the predicted mean calculated from the forecast. Also, for about 70-80\% of the hours, the realized generation is within the 10-90th quantile band, which is consistent with the coverage estimates shown in Table~\ref{tab: data_metric_summary}. 

For assets that are only weakly influenced by the training locations, such as Aguayo (which is relatively far from most training locations and hence is correlated to only a few of them, see the left panel of Figure~\ref{fig: model_scorr}), the predictive distribution has a mean $\hat{\mu}_{m^*,t,d^*}$ close to zero. Hence, the scenario mean is effectively the forecast $\Tilde{p}^{F}_{m^*,\cdot,d^*}$ plus $\bar{y}_{m^*}$, see the almost constant gap between the blue and green curves in the left and middle panels of Figure~\ref{fig: scenario_asset} (in Aguayo's case, the average forecast error $\bar{y}_{m^*}$ is highly positive).   
%
In contrast, for assets that are highly correlated with observed locations, such as Whirlwind (see right panel of Figure~\ref{fig: model_scorr}), more information is obtained from the observed data and $\hat{\mu}_{m^*,t,d^*}$ has a non-trivial shape. Thus, the difference between the scenario mean and forecast is influenced by both $\bar{y}_{m^*}$ and the predictive mean. Moreover, thanks to information fusion we observe notable improvement in predictive performance (scenarios closer to actuals compared to forecast)  as well as tighter predictive bands. 

\begin{figure}[!htb]
\begin{minipage}[b]{0.326\textwidth}
\includegraphics[width=0.95\textwidth]{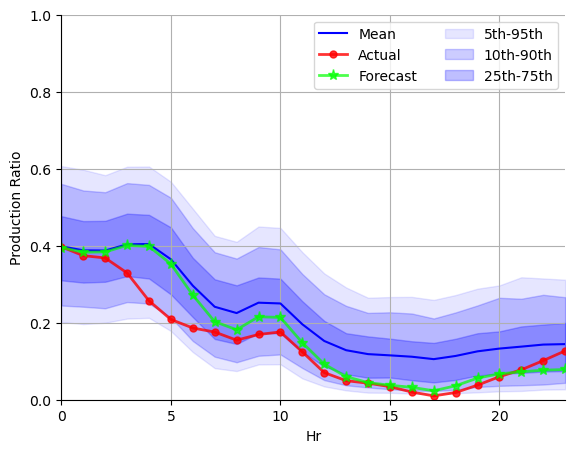}
\hspace*{30pt} Far West 02/07
  \end{minipage}
 \begin{minipage}[b]{0.326\textwidth}
\includegraphics[width=0.95\textwidth]{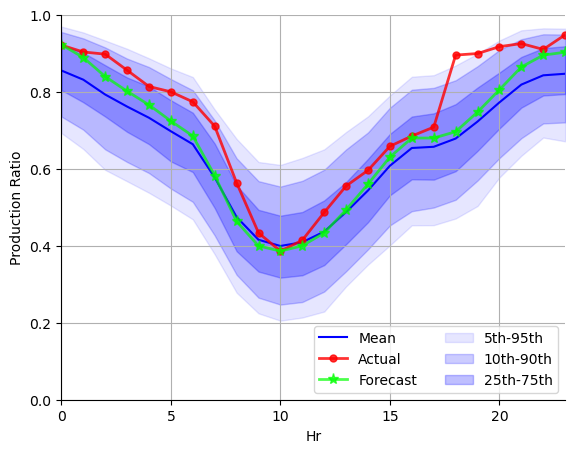}
\hspace*{40pt} Far West 04/12
  \end{minipage}
  \begin{minipage}[b]{0.326\textwidth}
\includegraphics[width=0.95\textwidth]{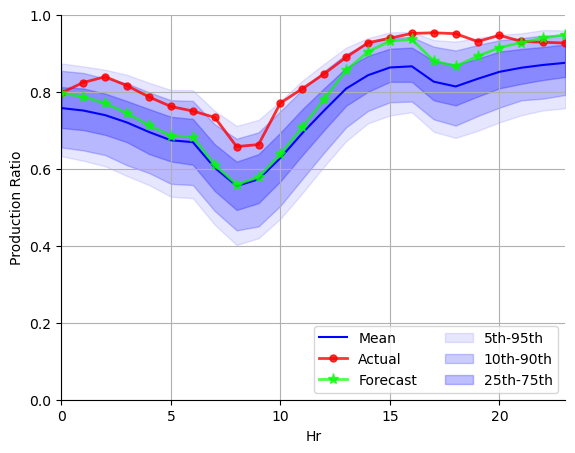}
\hspace*{40pt} North 04/12
  \end{minipage}
  \caption{Zone-level scenarios of wind power ratios simulated from the fitted M12-2-1 model. We show two different ERCOT zones on 2 representative days. 
  }
  \label{fig: scenario_region}
\end{figure}

An important use case of area-wide wind scenarios is bottom-up modeling of \emph{zonal} renewable generation which drives respective financial markets of zonal price indices and ancillary services. A joint probabilistic model of locational generation can be naturally summed across multiple assets to obtain consistent scenarios of aggregated production. As an illustration, Figure~\ref{fig: scenario_region} visualizes such simulations for the aggregated power ratios in the Far West and North zones of ERCOT. These simulations are obtained by first simulating individual assets like in Figure \ref{fig: scenario_asset} and then summing and re-normalizing in terms of zonal nameplate capacity across the given ERCOT zone. 

In Figure \ref{fig: scenario_region}, we sample scenarios from the joint distribution of asset wind power ratios given the day-ahead forecast only. After doing a weighted average to compute the zonal power ratio, the figure shows the mean of 1000 simulations, several respective quantile bands, as well as the 
hourly zonal forecast $\Tilde{p}^{F}$ and the actual generation ratio $\Tilde{p}^{A}$. Since the scenarios represent zonal generation based solely on the day-ahead forecast, the scenario mean is roughly the forecast plus the average $\bar{y}$ within the zone. Compared to the asset-level Figure~\ref{fig: scenario_asset}, we observe much tighter uncertainty bands and lower RMSEs at the zonal level thanks to spatial averaging. The
standard deviation of the zone-level scenario is about 10-12\% around the forecast, yielding an 80\% interval band  of approximately 0.2-0.3. 

\section{Conclusion}
\label{sec: conclusion}

In this paper we propose a probabilistic framework for spatiotemporal wind power generation modeling. Our model effectively integrates both spatial and temporal dependencies through employing GPs with data-driven input warping. 

The developed model contributes to the larger statistical toolbox supporting the integration of renewable resources in the power grid. The availability of probabilistic spatially-calibrated scenarios enables novel approaches to risk management and optimization. Access to the distribution of day-ahead generation, both locally and across regions of interest, allows better quantification of deviations from the forecast, critical for deploying sufficient (but not excessive) generating reserves, as well as optimization of economic dispatch throughout the day. Understanding the statistics of both asset-level and zonal-level forecast errors enables deeper analysis of likely congestion in the transmission network which in turn drives renewable curtailment. Ultimately, probabilistic scenarios are essential for designing new risk-aware unit commitment algorithms. All these applications are beyond the scope of our work but highlight how our contribution fits into a value chain of quantitative analysis for grid operators. 

Our algorithm was tailored to the specifics of the ARPA-E Texas-7k dataset. While it could certainly be applied to other similar datasets, several different extensions are warranted for other related contexts.  In our analysis, we modeled the discrepancy between the actual wind generation and its day-ahead forecasts. A natural extension would be to jointly model both the actual and forecasted generation through multi-output GPs \cite{de2021mogptk, liu2018remarks}. This would allow a better handle on the $[0,1]$ constraints of actual generation and more faithfully capture the forecast-dependent features (such as positive/negative bias when forecast is close to zero or one). 

A different direction for future work is to build a true temporal model that treats time as running continuously rather than as 24-hour chunks. This would be relevant for working with intra-day forecasts, where the operator wishes to condition on the morning realized generation to better quantify the uncertainty for the evening dispatch. Warped GPs would be a natural analogue to stochastic differential approaches for this purpose that have been proposed for single-asset settings \cite{tankov2022stochastic}. A third extension is to investigate alternative approaches to capturing spatial non-stationarity, such as deep GPs or autoregressive recurrent neural networks \cite{salinas2020deepar}.

\bibliography{ref}

\appendix
\renewcommand\thefigure{\thesection.\arabic{figure}}   
\renewcommand\thetable{\thesection.\arabic{table}} 
\section{Appendix}
\setcounter{figure}{0} 
\setcounter{table}{0}

\begin{figure}[!htb]
  \centering
  \begin{minipage}[b]{0.325\textwidth}
\includegraphics[width=0.9\textwidth]{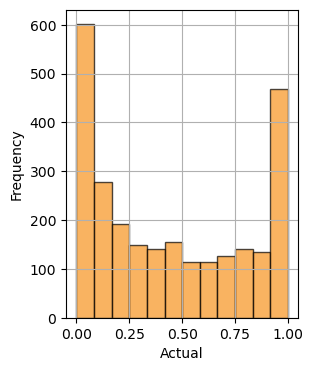}
  \end{minipage}  
  \begin{minipage}[t]{0.325\textwidth}
  \includegraphics[width=0.9\textwidth]{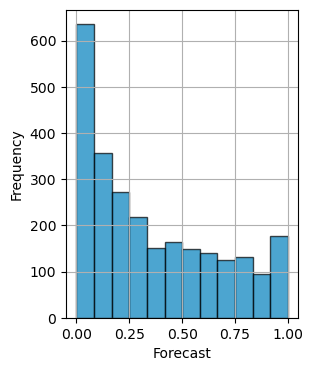} \end{minipage} 
  \begin{minipage}[t]{0.325\textwidth}
  \includegraphics[width=0.9\textwidth]{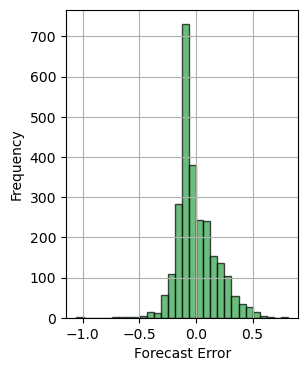} \end{minipage}
  \caption{Distribution of power ratios at Ajax Wind Farm. 
  Compared to the actual ($\Tilde{p}^{A}$) and forecast ($\Tilde{p}^{F}$) wind power ratios, the forecast error ($y$) is closer to a Gaussian distribution.
  } 
  \label{fig: ajax_histogram}
\end{figure}

\begin{table}[!htb]
    \small
    \centering
    \begin{tabular}{c c |c c c c c}\toprule
    \textbf{True/Model}
    & \textbf{Actual} 
    & \textbf{SE} 
    & \textbf{M52}
    & \textbf{M32}
    & \textbf{M12} \\
    \midrule
        \textbf{SE} 
        & \makecell[c]{$(2.00,1.00)$\\ $0.030$}
        & \makecell[c]{$(2.25, 1.03)$\\
        $0.035$}
        & \makecell[c]{$(2.26, 1.58)$\\
        $0.041$}  
        & \makecell[c]{$(2.08, 2.16)$\\$0.036$}
        & \makecell[c]{$(1.48, 5.37)$\\$0.015$}
        
        \\
        \midrule
        \textbf{M52} 
        & \makecell[c]{$(2.00,1.00)$\\ $0.030$} 
        & \makecell[c]{$(2.24, 0.65)$\\$0.031$} 
        & \makecell[c]{$(2.16, 1.02)$\\$0.033$} 
        & \makecell[c]{$(2.23,1.63)$\\ $0.043$} 
        & \makecell[c]{$(1.48, 4.03)$\\ $0.016$} 
        
        \\
        \midrule
        \textbf{M32} 
        & \makecell[c]{$(2.00,1.00)$ \\ $0.030$} 
        & \makecell[c]{$(2.24, 0.43)$\\$0.027$} 
        & \makecell[c]{$(2.16, 0.68)$\\$0.029$} 
        & \makecell[c]{$(2.11, 1.03)$\\ $0.033$} 
        & \makecell[c]{$(1.53, 3.32)$\\ $0.018$} 
        
        \\
        \midrule
        \textbf{M12} 
        & \makecell[c]{$(2.00,1.00)$\\ 0.030} 
        & \makecell[c]{$(2.42, 0.11)$\\ 0.031} 
        & \makecell[c]{$(2.41, 0.17)$ \\0.030} 
        & \makecell[c]{$(2.41, 0.24)$ \\ 0.031} 
        & \makecell[c]{$(1.99, 1.07)$ \\ 0.031}
        \\

        \bottomrule
    \end{tabular}
    \caption{Estimated kernel parameters, range $\hat{\bm{\rho}}=(\hat{\rho}_T, \hat{\rho}_S)$ and variance $\hat{\eta}$, for the synthetic experiment in Section \ref{sec: kernel_eval}.}
    \label{tab: ker_eval_params}
\end{table}

\begin{table}[!htb]
    \small
    \centering
    \begin{tabular}{c c|c c c c c}\toprule
    \textbf{True/Model}
    & \textbf{Actual Loss} 
    & \textbf{SE} 
    & \textbf{M52}
    & \textbf{M32}
    & \textbf{M12} \\
    \midrule
        \textbf{SE} 
        & \makecell[c]{$-1002.11$ \\ }
        & \makecell[c]{$\mathbf{-1002.10}$\\$\mathbf{0.558}$\\ $\mathbf{0.2279}$
        }
        & \makecell[c]{$-1001.94$ \\$0.487$\\ $0.2280$
        }  
        & \makecell[c]{$-1001.32$ \\$0.487$\\$0.2280$
        }
        & \makecell[c]{$-994.21$ \\$0.557$\\$0.2315$
        }
        
        \\
        \midrule
        \textbf{M52} 
        & \makecell[c]{$-997.13$ \\ } 
        & \makecell[c]{$-996.44$ \\$0.531$\\ $0.2285$
        } 
        & \makecell[c]{$\mathbf{-997.13}$ \\$0.540$\\ $\mathbf{0.2283}$
        } 
        & \makecell[c]{$-996.80$ \\$0.541$\\ $0.2286$
        } 
        & \makecell[c]{$-989.98$ \\$\mathbf{0.609}$\\ $0.2318$
        } 
        
        \\
        \midrule
        \textbf{M32} 
        & \makecell[c]{$-990.60$ \\ } 
        & \makecell[c]{$-987.64$ \\$0.541$\\$0.2307$
        } 
        & \makecell[c]{$-990.00$ \\$\mathbf{0.572}$\\$0.2302$
        } 
        & \makecell[c]{$\mathbf{-990.60}$ \\$0.449$\\ $\mathbf{0.2301}$
        } 
        & \makecell[c]{$-985.56$ \\$0.510$\\ $0.2330$
        } 
        
        \\
        \midrule
        \textbf{M12} 
        & \makecell[c]{$-959.94$ \\} 
        & \makecell[c]{$-943.38$ \\$0.077$\\ 0.3099
        } 
        & \makecell[c]{$-949.82$ \\$0.273$ \\ 0.2821
        } 
        & \makecell[c]{$-953.74$ \\$0.189$ \\ 0.2730
        } 
        & \makecell[c]{$\mathbf{-959.94}$ \\$\mathbf{0.549}$ \\ $\mathbf{0.2706}$
        }
        \\

        \bottomrule
    \end{tabular}
    \caption{Goodness of fit and predictive evaluation metrics for the synthetic experiment in Section \ref{sec: kernel_eval}. Training loss, Kolmogorov-Smirnov (KS) test p-values, and average interval score $\text{Avg IS}_{0.05}$. Lower loss, larger p-values and smaller interval scores indicate better performance. The lowest $\text{Avg IS}_{0.05}$ is typically achieved when $M_{\text{Model}}=M_{\text{True}}$. Notably, the other three kernels get lower p-values of KS test than in other cases when $M_{\text{True}}=M12$.}
    \label{tab: ker_eval_metric1}
\end{table}

\begin{table}[!htb]
    \centering
    \small
    \begin{tabular}{c|cccc|cccc}
        \toprule
        
        & \multicolumn{4}{c|}{\textbf{RMSE}} & \multicolumn{4}{c}{$\mathbf{C}_{0.2}$} \\
        \cmidrule{2-9}
        \textbf{True/Model}
        & \textbf{SE} & \textbf{M52} & \textbf{M32} & \textbf{M12} & \textbf{SE} & \textbf{M52} & \textbf{M32} & \textbf{M12}
        
        \\
        \midrule
        \textbf{SE} 
        & $\mathbf{0.04946}$
            & $0.04948$
            & $0.04953$
            & $0.04997$
              & $80.8$\%
            & $80.9\%$
            & $\mathbf{80.2\%}$
            & $81.7\%$
            \\
        \textbf{M52}
        & $\mathbf{0.04951}$
            & $0.04952$
            & $0.04957$
            & $0.05003$ 
            & $80.9\%$
            & $80.9\%$
            & $\mathbf{80.6\%}$
            & $81.5\%$
            \\
        \textbf{M32}
        & $0.04987$
            & $\mathbf{0.04980}$
            & $0.04981$
            & $0.05018$
           & $\mathbf{80.2\%}$
            & $80.3\%$
            & $80.7\%$
            & $81.9\%$  
            
            \\
        \textbf{M12}
        & $0.06169$
            & $0.05868$
            & $0.05770$
            & $\mathbf{0.05704}$
           & $77.5\%$
            & $78.2\%$
            & $\mathbf{78.8\%}$
            & $81.7\%$
            \\
        \bottomrule
    \end{tabular}
    \caption{Root Mean Squared Error (RMSE) and Coverage $C_{0.20}$ for the synthetic experiment in Section \ref{sec: kernel_eval}. Lower RMSE and closer to $80\%$ coverage means better performance. The lowest RMSE is typically observed when $M_{\text{Model}}=M_{\text{True}}$. The SE, M52, and M32 kernels exhibit similar coverage when $M_{\text{True}}$ is one of them, while they deviate from 80\% when $M_{\text{True}}=M12$.}
    \label{tab: ker_eval_metric2}
\end{table}

\begin{table}[!htb] 
    \centering
    \small
        \begin{tabular}{c | c c c c c c c}\toprule
    \textbf{Model}
    & \textbf{BIC} 
    & $\bm{\eta}$
    & $(\rho_T, \rho_S)$
    & $\sigma$
    & $\mathbf{w}$
    &  $\bm{\gamma}$
    & $a$
    
    \\
    \midrule
    \multicolumn{8}{c}{Single Warping Layer Run}  \\ \midrule
        True
        & / 
        & $0.030$
        & $(2.00,1.00)$
        & $0.05$
        & $(-0.70, 1.20)$
        & $(0.30, 0.60)$
        & $0.25$
        \\
        $W_1^{(M_1)}$
        & $-976.42$ 
        & $0.033$
        & $(2.11, 1.01)$
        & $0.05$
        & $(-0.66, 1.24)$
        & $(0.31, 0.61)$
        & $0.24$
        \\
        $W_1^{(M_0)}$ 
        & $-980.19$ 
        & $0.025$
        & $(2.08, 0.66)$
        & $0.05$ 
        & \multicolumn{3}{c}{N/A} 
        \\ \midrule 
\multicolumn{8}{c}{Double Warping Layer Run}  \\
    \midrule
        True
        & / 
        & $0.030$
        & $(2.00,1.00)$
        & $0.05$
        & \makecell[c]{$(1.2, 1.0)$\\$(-0.7, 1.5)$}
        & \makecell[c]{$(0.40, 0.60)$\\$(0.20,0.80)$}
        & \makecell[c]{$0.18$\\$0.25$}
        \\
        $W_2^{(M_2)}$
        & $-967.79$ 
        & $0.032$
        & $(2.11, 0.91)$
        & $0.05$
         & \makecell[c]{$(-0.71,  1.17)$\\$(-0.48,  0.58)$}
        & \makecell[c]{$(0.29, 0.65 )$\\$(0.37, 0.45)$}
        & \makecell[c]{0.16\\0.16}
        \\
        $W_2^{(M_1)}$
        & $-976.42$ 
        & $0.033$
        & $(2.11, 1.01)$
        & $0.05$
         & \makecell[c]{$(-0.66, 1.24)$}
        & \makecell[c]{$(0.31, 0.61)$}
        & \makecell[c]{$0.24$}
        \\
        $W_2^{(M_0)}$ 
        & $-980.19$ 
        & $0.025$
        & $(2.08,0.66)$
        & $0.05$ & \multicolumn{3}{c}{N/A} 
        \\
         \bottomrule
    \end{tabular}
\caption{Summary of the true setting and training results from three models for the experiment in Section \ref{sec: warp_sim}. For the single-layer case study, both models have similar BIC values. $W_1^{(M_1)}$ successfully recovers the kernel and warping hyperparameter values.  For the two-layer case study, $W_2^{(M_0)}$ shows a larger BIC and inaccurate kernel parameters estimation. None of the three models recover the warped space.}
\label{tab: W2_train_tables}
\end{table}

\begin{figure}[!htb]
  \centering
\includegraphics[width=0.5\textwidth]{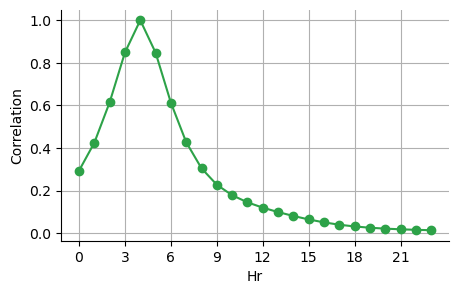}
\caption{Visualizing estimated temporal correlation correlation structure of the fitted M12-2-1 model in Section \ref{sec: data_analysis}, relative to 4 AM.}
\label{fig: model_tcorr}
\end{figure}

\begin{figure}[!htb]
  \centering
  \begin{tabular}{cc}
\includegraphics[width=0.465\textwidth]{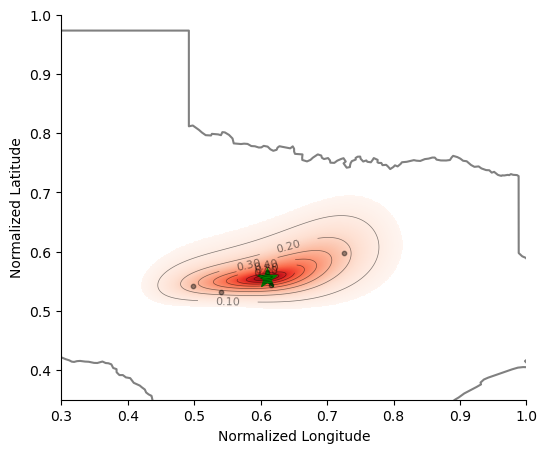} & 
\includegraphics[width=0.465\textwidth]{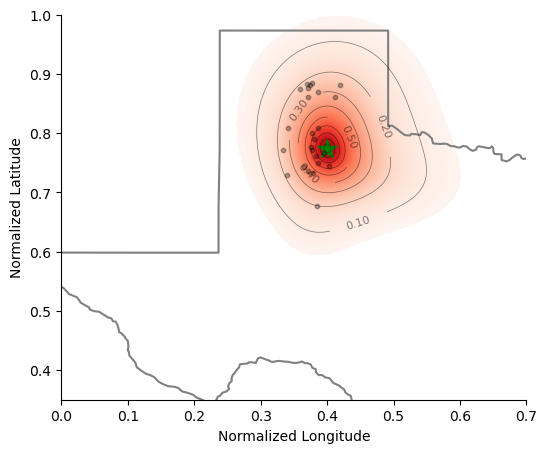} \\
$\qquad$ Aguayo Wind & $\qquad$ Whirlwind \end{tabular}

  \caption{Visualizing spatial correlation structure of the fitted M12-2-1 model in Section \ref{sec: data_analysis}, showing the relationship between a specific wind farm, marked by the green star, and neighboring areas in the ERCOT region. Black dots indicate observed assets with spatial correlations greater than 0.2 with the specified wind farm.
  Left: Estimated spatial correlation for Aguayo wind farm. Right: Estimated spatial correlation for Whirlwind.
  The non-elliptical patterns are due to the input warping $g(\cdot)$. Correlations at the indicated reference inputs are 1 by definition. Correlation values below 0.05 are masked in white.}
  \label{fig: model_scorr}
\end{figure}

\end{document}